\documentclass[11pt]{article}
\usepackage[table]{xcolor}
\usepackage[final]{acl}

\usepackage{times}
\usepackage{latexsym}
\usepackage{amsmath}
\usepackage{enumitem}
\usepackage{amssymb}
\usepackage{hyperref}
\usepackage{algorithm}
\usepackage{algorithmic}
\usepackage{amsmath}
\usepackage{tikz}
\usepackage{float}
\usepackage{pdfpages}
\usepackage[utf8]{inputenc}
\usepackage{xcolor}
\usepackage{tcolorbox}
\tcbuselibrary{skins, breakable}
\usepackage{listings}
\usepackage{amssymb}
\usepackage{tabularx}
\usepackage{orcidlink}
\usepackage{adjustbox}
\usetikzlibrary{shapes.geometric, arrows.meta, positioning, fit, backgrounds, shapes.misc}

\usepackage{booktabs} 
\definecolor{usercolor}{RGB}{66, 135, 245}   %
\definecolor{llmcolor}{RGB}{52, 168, 83}     %
\definecolor{z3color}{RGB}{241, 196, 15}     %
\definecolor{errorcolor}{RGB}{231, 76, 60}   %
\definecolor{successcolor}{RGB}{46, 204, 113} %
\definecolor{bg_iter1}{RGB}{253, 242, 242}   %
\definecolor{bg_iter2}{RGB}{240, 255, 240}   %
\def\checkmark{\tikz\fill[scale=0.4](0,.35) -- (.25,0) -- (1,.7) -- (.25,.15) -- cycle;}

\definecolor{lightblue}{RGB}{173, 216, 230}
\definecolor{beige}{RGB}{245, 222, 179}
\definecolor{lightgreen}{RGB}{144, 238, 144}
\definecolor{lightyellow}{RGB}{255, 255, 153}
\definecolor{lightpink}{RGB}{255, 192, 203}
\definecolor{darkteal}{RGB}{0, 128, 128}

\tikzset{
    startstop/.style={rounded rectangle, minimum width=2cm, minimum height=0.8cm, 
                      text centered, draw=black, fill=lightblue, font=\tiny, thick},
    process/.style={rectangle, minimum width=2cm, minimum height=0.8cm, 
                   text centered, draw=black, fill=lightblue, font=\tiny, thick},
    entity/.style={rectangle, minimum width=2cm, minimum height=0.8cm, 
                  text centered, draw=black, fill=beige, font=\tiny, thick},
    claim/.style={rectangle, minimum width=2cm, minimum height=0.8cm, 
                 text centered, draw=black, fill=lightgreen, font=\tiny, thick, rounded corners=2pt},
    decision/.style={diamond, minimum width=1.8cm, minimum height=1.8cm, 
                    text centered, draw=black, fill=lightyellow, font=\tiny, aspect=1.5, thick},
    verification/.style={rectangle, minimum width=1.8cm, minimum height=0.7cm, 
                        text centered, draw=black, fill=lightpink, font=\tiny, thick, rounded corners=2pt},
    verifdiamond/.style={diamond, minimum width=1.6cm, minimum height=1.6cm, 
                        text centered, draw=black, fill=lightpink, font=\tiny, aspect=1.5, thick},
    acceptance/.style={rectangle, minimum width=2cm, minimum height=0.8cm, 
                      text centered, draw=black, fill=darkteal, text=white, font=\tiny, thick},
    acceptdec/.style={diamond, minimum width=2cm, minimum height=2cm, 
                     text centered, draw=black, fill=darkteal, text=white, font=\tiny, aspect=1.5, thick},
    arrow/.style={thick,->,>=Stealth}
}

\usepackage[T1]{fontenc}
\usepackage[utf8]{inputenc}
\usepackage{microtype}
\usepackage{soul}
\usepackage{stmaryrd}
\usepackage{inconsolata}
\usepackage{graphicx}
\usepackage{multirow}
\usepackage{multicol}
\usepackage{xcolor}
\usepackage{tcolorbox}
\tcbuselibrary{skins, breakable, listings}
\usepackage{booktabs}
\usepackage{tabularx}
\usepackage{colortbl}
\usepackage{enumitem}
\usepackage{amssymb}
\usepackage{pifont}
\usepackage{listings}
\usepackage{float}
\usepackage{longtable}

\definecolor{passgreen}{RGB}{34, 139, 34}
\definecolor{passbg}{RGB}{240, 248, 240}
\definecolor{failred}{RGB}{178, 34, 34}
\definecolor{failbg}{RGB}{255, 240, 240}
\definecolor{warnorange}{RGB}{255, 140, 0}
\definecolor{warnbg}{RGB}{255, 248, 225}
\definecolor{infoblue}{RGB}{0, 100, 180}
\definecolor{infobg}{RGB}{235, 245, 255}
\definecolor{codebg}{RGB}{250, 250, 250}
\definecolor{darkgray}{RGB}{60, 60, 60}
\definecolor{neutralgray}{RGB}{120, 120, 120}

\lstdefinestyle{smtstyle}{
    backgroundcolor=\color{codebg},
    basicstyle=\ttfamily\scriptsize\color{darkgray},
    keywordstyle=\color{blue}\bfseries,
    commentstyle=\color{gray}\itshape,
    stringstyle=\color{passgreen},
    morekeywords={assert, declare-fun, declare-const, declare-sort, define-fun, check-sat, get-unsat-core, forall, exists, and, or, not, =>, distinct, ite},
    frame=single,
    rulecolor=\color{lightgray},
    breaklines=true,
    captionpos=b,
    language=Lisp
}

\newcommand{\statusbadge}[2]{%
    \tcbox[on line, arc=2pt, colback=#1!10, colframe=#1, boxrule=0.5pt, 
    left=2pt, right=2pt, top=1pt, bottom=1pt, fontupper=\bfseries\scriptsize\color{#1}]{#2}%
}

\newcolumntype{Y}{>{\centering\arraybackslash}X}
\newcolumntype{L}{>{\raggedright\arraybackslash}X}

\usepackage{xspace}
\newcommand{\forge}{\textsc{VERGE}\xspace}

\title{\forge: Formal Refinement and Guidance Engine for Verifiable LLM Reasoning}

\author{
\textbf{Vikash Singh\textsuperscript{1}\thanks{Work done during internship at Amazon Web Services}},
\textbf{Darion Cassel\textsuperscript{2}},
\textbf{Nathaniel Weir\textsuperscript{2}}, 
\textbf{Nick Feng\textsuperscript{2}}, 
\textbf{Sam Bayless\textsuperscript{2}}
\\
\textsuperscript{1}Case Western Reserve University 
\textsuperscript{2}Amazon Web Services
\\
\texttt{vikash@case.edu,\{darionfc, nweir, nicfeng, sabayles\}@amazon.com}
}

\begin{document}
\maketitle

\begin{abstract}
Despite the syntactic fluency of Large Language Models (LLMs), ensuring their logical correctness in high-stakes domains remains a fundamental challenge.
We present a neurosymbolic framework that combines LLMs with SMT solvers to produce verification-guided answers through iterative refinement. Our approach decomposes LLM outputs into atomic claims, autoformalizes them into first-order logic, and verifies their logical consistency using automated theorem proving. We introduce three key innovations: (1) multi-model consensus via formal semantic equivalence checking to ensure logic-level alignment between candidates, eliminating the syntactic bias of surface-form metrics,  (2) semantic routing that directs different claim types to appropriate verification strategies: symbolic solvers for logical claims and LLM ensembles for commonsense reasoning, and (3) precise logical error localization via Minimal Correction Subsets (MCS), which pinpoint the exact subset of claims to revise, transforming binary failure signals into actionable feedback. Our framework classifies claims by their logical status and aggregates multiple verification signals into a unified score with variance-based penalty. The system iteratively refines answers using structured feedback until acceptance criteria are met or convergence is achieved. This hybrid approach delivers formal guarantees where possible and consensus verification elsewhere, advancing trustworthy AI. With the GPT-OSS-120B model, \forge demonstrates an average performance uplift of 18.7\% at convergence across a set of reasoning benchmarks compared to single-pass approaches.

\end{abstract}
\section{Introduction}
\label{sec:introduction}

\begin{figure}[t]
\includegraphics[width=0.5\textwidth]{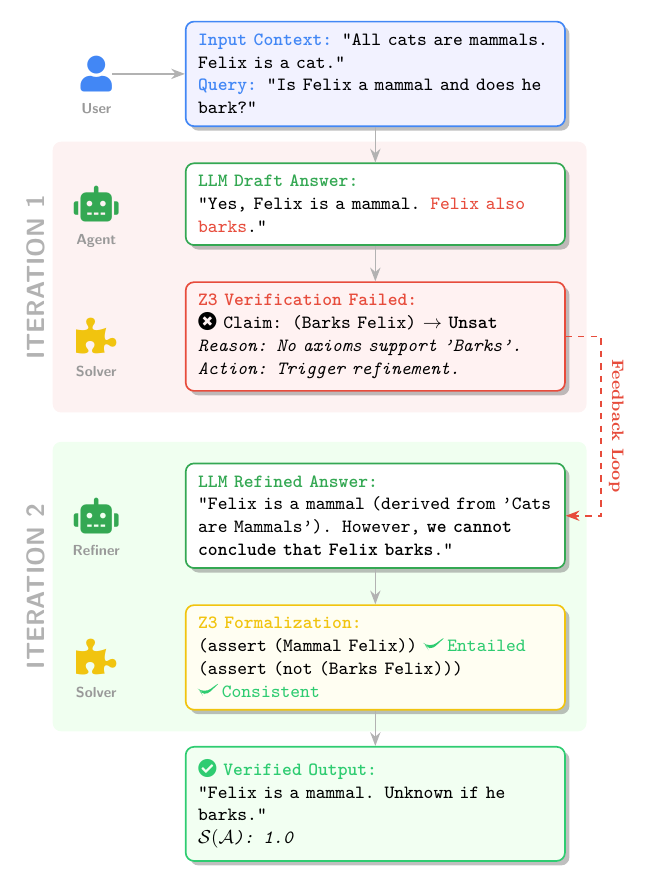}
\caption{\forge correcting LLM Hallucinations via Formal Verification. The {\color{yellow!70!black}solver} detects an unsupported {\color{red}claim} and guides the LLM to a consistent {\color{green!70!black}answer} through MCS-based feedback.}
\label{fig:example}
\end{figure}

Large Language Models (LLMs) have demonstrated remarkable capabilities across diverse reasoning tasks, from mathematical problem-solving \citep{lewkowycz2022solving,hendrycks2021measuring} to code generation \citep{chen2021evaluating,austin2021program} and logical inference \citep{tafjord2021proofwriter}. Despite these advances, ensuring the correctness of LLM-generated answers remains a critical barrier in high-stakes domains such as legal policy compliances, healthcare and finance etc. While recent models achieve impressive benchmark performance \citep{openai2025o4mini,anthropic2025opus4_5}, they rely on statistical likelihood maximization \citep{ouyang2022training} rather than logical deduction. Consequently, they operate without mechanisms for provable correctness, making them prone to hallucinations and internal contradictions.

Current verification strategies such as self-consistency \citep{wang2022self}, process supervision \citep{lightman2023verify}, and self-refinement \citep{madaan2023self,shinn2023reflexion} provide \textit{heuristic} rather than \textit{formal} guarantees. Even multi-agent debate frameworks \citep{du2023improving,liang2023encouraging} merely achieve consensus, which does not imply correctness. To achieve verifiable reasoning, neuro-symbolic methods \citep{ganguly2025grammars,ganguly2024proof,feng2025vericot,pan2023logiclmempoweringlargelanguage,callewaert2025verus,olausson2023linc,garcez2019neural,mao2019neuro} and semantic parsing \citep{he2025star,mcginness2024automated,zettlemoyer2005learning,dong2016language} have attempted to bridge natural language with formal logic. However, these approaches face a fundamental \emph{semantic gap}: natural language is inherently ambiguous, and rigid formalization often fails on open-domain claims \citep{church1936unsolvable,turing1936computable}.

We present \textbf{\forge}, a framework that mitigates this gap by combining LLMs with Satisfiability Modulo Theories (SMT) solvers \citep{barrett2009satisfiability,demoura2008z3} to produce verification-guided answers with formal guarantees for logical/mathematical claims through iterative refinement, as illustrated in Figure~\ref{fig:example}. Unlike standard feedback loops, \forge leverages the SMT solver's ability to extract \textit{unsatisfied assertions} \citep{zhang2003extracting,nadel2014efficient}. This allows us to compute \textbf{Minimal Correction Subsets (MCS)} \citep{marques2013computing}, identifying a minimal set of modifications to the atomic claims  sufficient to restore consistency and move from probabilistic self-correction to logical self-consistency correction.%

\paragraph{The Expressivity Trade-off and Pragmatic Verification.}A key insight of our work is that enforcing formal verification on all claims is fundamentally misaligned with the ambiguity of natural language. Rather than attempting to bridge this gap universally, a theoretically intractable goal, we adopt a \textbf{pragmatic stance}: Apply formal verification where the semantic gap is narrow (mathematical/logical claims), and fall back to consensus-based verification where it is wide (commonsense/vague claims). \forge introduces a verification cascade with semantic routing to implement this strategy. This hybrid approach provides formal guarantees for a verifiable subset of claims while maintaining the system's ability to handle broad, real-world reasoning tasks.

Our work introduces:
\begin{enumerate}[nosep, leftmargin=*]
   \item  \textbf{High-Fidelity Consensus via SMT:} We bridge the semantic gap by enforcing semantic equivalence ($\phi_a \iff \phi_b$) among candidate formalizations. Unlike syntactic metrics (e.g., BLEU, Jaccard) which fail on variable renaming or structural permutation, we utilize the solver to prove that different candidate formulas yield identical truth tables, ensuring robust consensus.
    \item \textbf{Actionable Feedback via MCS:} We use Minimal Correction Subsets (MCS) \citep{marques2013computing} to provide specific, logically derived feedback (e.g., "claim $C_2$ does not hold") rather than generic error signals.
    \item \textbf{Flexible Neuro-symbolic Integration:} A semantic routing framework that balances the precision of SMT solvers with the flexibility of LLMs, avoiding the pitfalls of forcing undecidable language into decidable theories.
\end{enumerate}

\section{Related Work}
\label{sec:related}
\paragraph{Probabilistic vs. Formal Verification.}
Standard LLM reasoning strategies rely on \textit{probabilistic} confidence. Methods like self-consistency \citep{wang2022self} and process supervision \citep{lightman2023verify,uesato2022solving} aggregate samples or train verifiers on human labels, but cannot guarantee logical soundness. Self-refinement approaches \citep{madaan2023self,shinn2023reflexion,welleck2022generating} use the model to critique itself, often failing due to the faithfulness gap where reasoning does not match output \citep{lyu2023faithful,huang2024large}. Multi-agent debate \citep{du2023improving,liang2023encouraging} achieves consensus, not truth, recent work shows self-correction can even degrade performance \citep{huang2024large,kamoi2024can}. In contrast, \forge uses SMT solvers \citep{barrett2009satisfiability,demoura2008z3} to provide \textit{mathematically proven} feedback. Unlike tool-augmented LLMs \citep{schick2023toolformer,gou2024critic} that use tools for execution (e.g., calculators), we use tools for \textit{consistency checking}, computing MCS. \citep{zhang2003extracting,nadel2014efficient} to identify exactly which premises contradict the generated answer.%
\paragraph{Neuro-Symbolic Integration and the Semantic Gap.}
Traditional semantic parsing \citep{dong2016language,berant2013semantic,zettlemoyer2005learning,xuadaptive2026,planninghao2025} maps language to executable logical forms but requires expensive supervision. Recent work extends this to theorem proving \citep{polu2020generative,polu2022formal,jiang2022draft,azerbayev2023proofnet} and augmenting LLMs and VLMs with symbolic solvers \citep{pan2023logiclmempoweringlargelanguage,olausson2023linc,callewaert2025verus,singh2026guaranteesclinicalreasoningvision,feng2025vericot}, but these require fully formalizable domains. Prior neuro-symbolic integration \citep{singh2026guaranteesclinicalreasoningvision,garcez2019neural,kautz2020third,hall2026neurosymbolicauditingnaturallanguagesoftware} and grammar-based approaches \citep{ganguly2025grammars,ganguly2024proof} struggle with the semantic gap \citep{church1936unsolvable,turing1936computable} the mismatch between ambiguous natural language and rigid formal systems\citep{chen2026learninggenerateformallyverifiable}. \forge targets \textit{open-domain} natural language where full formalization is often impossible. We introduce \textbf{semantic routing}, rather than forcing vague or commonsense claims into rigid first-order logic, we route them to a consensus-based soft verifier. This treats the semantic gap as an inherent property of language requiring hybrid verification.
\paragraph{Automated Reasoning for Repair.}
Our feedback mechanism adapts MCS computation \citep{marques2013computing,liffiton2009algorithms} from constraint programming to NLP. MCS identifies a minimal set of constraints to relax to restore satisfiability. Recent work applies MCS to constraint relaxation \citep{bacchus2015finding} and automated debugging \citep{morgado2013iterative}. We innovate by translating MCS output into natural language feedback, guiding the LLM to \textit{rewrite} specific atomic claims. This prioritizes interpretability and convergence speed ($O(m \times \text{SAT})$ greedy approximation) over theoretical optimality. To our knowledge, \forge is the first to apply MCS-based feedback to guide iterative refinement in LLM reasoning.

\begin{figure*}
    \centering
    \includegraphics[width=\textwidth]{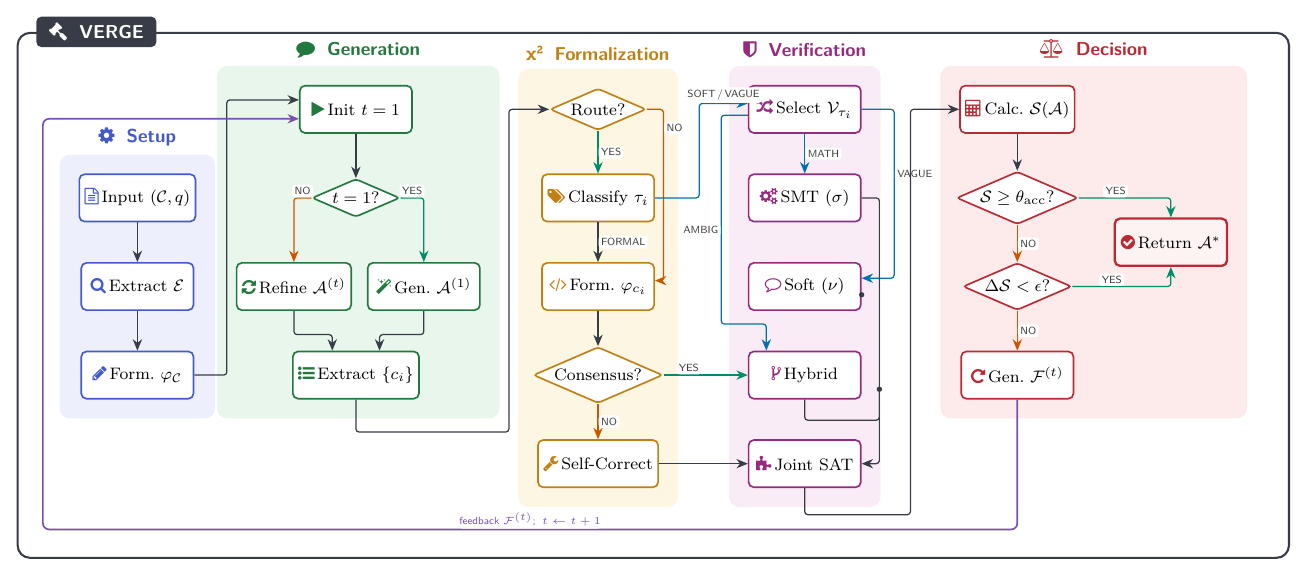}
    \caption{Overview of \textbf{\forge}: The pipeline is structured into five distinct stages: {\color{blue!40}Setup} prepares the context $\mathcal{C}$ and entities $\mathcal{E}$; {\color{green!40!black}Generation} produces and refines answers $\mathcal{A}^{(t)}$ iteratively; {\color{orange!40}Formalization} classifies claim types $\tau_i$ and generates SMT formulas $\varphi_{c_i}$; {\color{purple!40}Verification} routes claims to SMT, Soft, or Hybrid verifiers based on semantic type; and {\color{red!40}Decision} computes the aggregate score $\mathcal{S}(\mathcal{A})$ to either accept the answer $\mathcal{A}^*$ or generate feedback $\mathcal{F}^{(t)}$ for the next iteration.} 
    \label{fig:overview}
\end{figure*}

\section{Methodology}
\label{sec:methodology}
\textbf{Problem Formulation.} Given a context $\mathcal{C} = \{p_1, \ldots, p_m\}$ of premise statements and a query $q$, we aim to produce a verified answer $\mathcal{A}^*$ composed of atomic claims $\{c_1, \ldots, c_n\}$ with maximal verification coverage. Here, $\mathcal{A}^*$ is a refined version of the candidate answer $\mathcal{A}$ obtained through verification. We formalize this as maximizing the verification score $\mathcal{S}(\mathcal{A}) \in [0,1]$ subject to two constraints for each claim $c_i$: (1) \textit{Consistency}: $\text{SAT}(\varphi_{\mathcal{C}} \wedge \varphi_{c_i}) = \text{true}$, and (2) \textit{Entailment}: $\text{SAT}(\varphi_{\mathcal{C}} \wedge \neg \varphi_{c_i}) = \text{false}$, where $\varphi$ denotes the logical formalization function that maps natural language statements to SMT constraints. The consistency constraint verifies that each claim is compatible with the context, while the entailment constraint ensures that each claim is a logical consequence of the context. These constraints provide formal equivalence guarantees where logic permits, while falling back to semantic consistency (soft verification, see \S\ref{subsec:verification}) otherwise.

\textbf{Pipeline.} The pipeline (Fig.~\ref{fig:overview}) executes iteratively: (1) entity extraction, (2) generation, (3) decomposition, (4) formalization \& verification, and (5) refinement.

\subsection{Entity Extraction and Generation}
\label{subsec:generation}
We first extract entities $\mathcal{E} = \text{Extract}(\mathcal{C}, q)$ (e.g., `Felix', `Monday', `Process A') to serve as typed  constants in SMT. At iteration $t$, we generate answer $\mathcal{A}^{(t)}$ via a language model $M$:
\begin{equation}
\label{eq:answer-generation}
\mathcal{A}^{(t)} = M(\mathcal{C}, q, \mathcal{A}^{(t-1)}, \mathcal{F}^{(t-1)})
\end{equation}
where $\mathcal{F}^{(t-1)}$ is structured feedback (see \S\ref{subsec:refinement}). The initial iteration ($t=1$) utilizes zero-shot prompting (e.g., $A^0 = M(\mathcal{C}, q, \emptyset, \emptyset)$).

\subsection{Claim Decomposition and Classification}
\label{subsec:decomposition}
We decompose $\mathcal{A}$ into atomic claims $\{c_1, \ldots, c_n\}$. To ensure our system is honest about what it can and cannot formally prove, we classify each claim $c_i$ into a semantic type $\tau_i \in \mathcal{T}$ via $M$:
\begin{equation}
\tau_i = \arg\max_{\tau \in \{\tau_{\text{M}}, \tau_{\text{L}}, \tau_{\text{T}}, \tau_{\text{P}}, \tau_{\text{C}}, \tau_{\text{V}}\}} P_{M}(\tau \mid c_i)
\end{equation}
where types correspond to \textit{Mathematical, Logical, Temporal, Probabilistic, Commonsense, and Vague} claims. These categories align with standard distinctions in semantic parsing to differentiate verifiable facts from subjective or probabilistic statements. This classification is crucial for minimizing "false formalization", the error of forcing ambiguous natural language into rigid logic. Vague claims are identified by the model as containing subjective predicates (e.g., "likely", "possibly") that lack binary truth values, preventing brittle SMT assertions.%

\subsection{SMT Formalization with Consensus}
\label{subsec:formalization}
For claims classified as logical or mathematical, we target the \textbf{QF\_UF} (Quantifier-Free Uninterpreted Functions) and \textbf{QF\_LIA} (Quantifier-Free Linear Integer Arithmetic) fragments within the SMT-LIB2 standard. Given the context $\mathcal{C}$ and query $q$, we first extract entities $\mathcal{E}$ as a set of typed constants. For example, from the statement ``All humans above age 18 are adults'', we extract entities $age$ of type $\mathbb{Z}$ and $adult$ of type AGE\_GROUP, which are declared as uninterpreted constants of the corresponding sorts. We then use generate candidate formulas $\phi = M(c_i, \mathcal{E})$ for each claim $c_i$ over the vocabulary in $\mathcal{E}$. To mitigate the stochastic nature of autoformalization, we generate $K=3$ candidate formulas $\{\phi_1, \ldots, \phi_K\}$ and check for consensus. Instead of relying on brittle syntactic overlap, we compute consensus based on semantic equivalence (see Appendix~\ref{app:semantic_equivalence} for the formal definition). Two formulas $\phi_a$ and $\phi_b$ are deemed equivalent if and only if their bidirectional entailment is valid, which we verify by querying the SMT solver:
\begin{equation}
\label{eq:equivalence}
\text{Equiv}(\phi_a, \phi_b) \iff \text{UNSAT}\left( \Sigma_{\mathcal{E}} \wedge \neg (\phi_a \leftrightarrow \phi_b) \right)
\end{equation}
where $\Sigma_{\mathcal{E}}$ represents the declarations of constants and uninterpreted functions derived from the entity extraction phase. %
This procedure constructs a semantic equivalence graph where edges represent proven logical identity.

A formalization is accepted only if a \textbf{majority consensus} is reached (i.e., the size of the largest semantic equivalence clique is $\ge \lceil K/2 \rceil$). Additionally, we employ \textbf{Round-Trip Translation} (SMT $\to$ Natural Language) as a semantic sanity check to ensure alignment with the source text. 
If consensus fails or the confidence (derived from clique size and round-trip similarity) is low, we trigger a self-correction step $\varphi^{\text{new}} = M(\varphi, \dots)$, constrained such that $\varphi^{\text{new}} \models \varphi$. This ensures the system strictly strengthens (and never weakens) the constraints during refinement. The strengthening constraint is verified by checking the unsatisfiability of $\varphi^{\text{new}} \wedge \neg \varphi$. To ensure decidability, we perform this check over the finite domain  of entities $\mathcal{E}$ extracted in the setup phase, effectively reducing the check to propositional logic or quantifier-free first-order logic(QF-FOL).

\subsection{Verification Cascade}
\label{subsec:verification}
We employ a hybrid strategy that routes claims to the most rigorous verifier available for their type, prioritizing formal guarantees where applicable.%
\paragraph{Semantic Routing (Flexibility Mechanism).}
We define a routing function $\mathcal{V}_{\tau_i}$:
\begin{equation}
\mathcal{V}_{\tau_i} = \begin{cases} 
\text{\textit{SMT-Verify}} & \text{if } \tau_i \in \{\tau_{\text{M}}, \tau_{\text{L}}, \tau_{\text{T}}\} \\
\text{\textit{Soft-Verify}} & \text{if } \tau_i \in \{\tau_{\text{C}}, \tau_{\text{V}}, \tau_{\text{P}}\} \\
\text{\textit{Hybrid}} & \text{else (or SMT error)}
\end{cases}
\end{equation}
While SMT provides provable correctness, it cannot natively handle vague predicates (e.g., `is similar to') without excessive and fragile axioms. By routing these to Soft-Verify, \forge preserves logical rigor where possible while preventing the system from falsely treating probabilistic or ambiguous reasoning as logical certainty.
\paragraph{SMT-Based Verification.} For logic-amenable claims, we check satisfiability using the Z3 solver. We assign statuses based on rigorous logical tests:
\begin{itemize}[itemsep=-10pt, parsep=0pt, topsep=0pt]
 \item {\textbf{Contradictory ($\sigma_{\text{C}}$):}} $\text{SAT}(\varphi_{\mathcal{C}} \wedge \varphi_{c_i}) = \text{False}$. The claim violates the context.\\
\item{\textbf{Entailed ($\sigma_{\text{E}}$):}} $\text{SAT}(\varphi_{\mathcal{C}} \wedge \neg\varphi_{c_i}) = \text{False}$. The claim is proven true (proof by contradiction).\\
\item{\textbf{Possible ($\sigma_{\text{P}}$):}} Consistent ($\text{SAT}(\varphi_{\mathcal{C}} \wedge \varphi_{c_i}) = \text{True}$) but not entailed. The context allows the claim but does not force it.\\
 \item{\textbf{Unknown ($\sigma_{\text{U}}$):}} Solver timeout or execution error.%
\end{itemize}
\textbf{Minimal Correction Sets (MCS).} 
For contradictions ($\sigma_{\text{C}}$), we compute the MCS to generate precise feedback. Let $\varphi_{c_i}$ be a set of clauses, the subset $S \subset  of \varphi_{c_i}$ is a minimal correction set (MCS) if 

\begin{itemize}[nosep]
\item SAT($\varphi_{\mathcal{C}}\wedge\varphi_{c_i} \setminus S$) = true 
\item $\forall S' \subset S$, SAT($  \varphi_{\mathcal{C}}\wedge\varphi_{c_i} \setminus S'$) = false
\end{itemize}

Intuitively, an MCS is a small subset of clauses that, if removed, will restore satisfiability. This provides actionable guidance to address contradictions (e.g., "remove constraint X")  rather than generic error messages.

\paragraph{Soft Verification.}\label{sec:soft-verification}
For claims unsuitable for SMT, we use an ensemble of LLM judges. We compute a confidence weighted majority vote (using self-reported confidence of the judge) by defining verdicts $v\in\{\text{Supported, Plausible, Unsupported, Uncertain}\}$.

\textbf{\textit{Constraint:}} To penalize the lack of formal guarantees, soft-verified claims are capped at a lower maximum score contribution than SMT-verified claims (see \S\ref{subsec:scoring}).
\paragraph{Hybrid Verification.} The Hybrid strategy acts as a robustness fallback. Claims routed to SMT that fail due to non-logical errors (e.g., syntax errors, undeclared variables, or timeouts) are automatically re-routed to Soft Verification. This prevents the pipeline from stalling on "Unknown" ($\sigma_U$) statuses due to correctable formalization issues, allowing the system to degrade gracefully from formal proof to probabilistic consensus.

\subsection{Score Aggregation}
\label{subsec:scoring}
We compute an aggregated verification score for the entire answer by combining verification results from both soft and hard verification across all atomic claims.
Let $\mathcal{A}$ be an answer to query $q$ given context $\mathcal{C}$. Suppose $\mathcal{A}$ is decomposed into atomic claims $c_1, \ldots, c_n$ and verified against $q$ and $\mathcal{C}$ to produce verification results $\sigma_1, \ldots, \sigma_n$, respectively. 
The verification score $S(\sigma_i)$ for each result is defined as:
\begin{equation}
    S(\sigma_i) = \begin{cases} 
1.0 & \text{if \textbf{Entailed}($\sigma_i$)}  \\
0.9 & \text{if \textbf{Supported}($\sigma_i$)} \\
0.7 & \text{if \textbf{Possible}($\sigma_i$)}  \\
0.0 & \text{if \textbf{Contradictory}($\sigma_i$) or else} 
\end{cases}
\end{equation}

These weights reflect verification rigor: formally entailed claims receive the maximum score (1.0), while soft-verified supported claims receive 0.9, ensuring the system favors provable logic over semantic consistency.
The final aggregated score $\mathcal{S}(\mathcal{A})$ integrates a variance-based penalty (Eq.~\ref{final-score}) to discourage ``gaming'' the system, where a model might generate claims that are individually confident but mutually contradictory under joint verification:
\begin{equation}
\label{final-score}
\mathcal{S}(\mathcal{A}) = \bar{S} \cdot \max\left(0.5, 1.0 - \frac{\sigma_S}{\bar{S} + 0.01}\right)
\end{equation}
where $\bar{S}$ is the mean of the verification scores $\{S(\sigma_1), \ldots, S(\sigma_n)\}$ and $\sigma_S$ is their standard deviation.

\paragraph{Iterative Refinement.}
\label{subsec:refinement}
At each iteration $t$, we generate feedback $\mathcal{F}^{(t)}$ containing:
(1) \textbf{Unsat Cores \& MCS} for contradictions, pinpointing exact logical conflicts;
(2) \textbf{Joint Conflicts} for mutually incompatible claims; and
(3) \textbf{Formalization Alerts} for low-confidence mappings.
The process repeats until $\mathcal{S}(\mathcal{A}) \geq 0.75$ and \texttt{JointSAT=True} (where \texttt{JointSAT} is the boolean result of the joint satisfiability check), or until convergence ($\Delta \mathcal{S} < 0.01$). \textbf{For Joint Consistency}, soft-verified claims are treated as atomic boolean variables ($b_i$) within the SMT solver. To capture interactions between soft and hard claims, the context formalization $\varphi_{\mathcal{C}}$ includes \textbf{bridging axioms} generated by the LLM (e.g., assertions linking vague predicates like ``small'' to numerical bounds). This allows the solver to detect if a mathematically proven claim ($\tau_{M}$) contradicts a commonsense claim ($\tau_{C}$) (e.g., Mathematical Claim ``$X > 10$'' vs Commonsense Claim ``$X$ is a small single-digit number'' which implies $X < 10$), ensuring holistic consistency even without full formalization of the commonsense component. If claims are not jointly consistent, we compute the MCS over the claims as refinement feedback, signaling a minimal patch to restore joint consistency.
Algorithm~\ref{alg:refinement} (Appendix) details the complete refinement procedure.

\section{Results}

\begin{table*}[ht!]
        \centering
        \footnotesize
\caption{\textbf{Comprehensive Performance Analysis.}  Results are averaged over 5 independent runs, with error bars (subscript) indicating standard deviation. We compare \forge against standard prompting (CoT, SC, SR) and neuro-symbolic baselines (DSB, LogicLM, LINC, PoT) across three different backbone models: GPT-OSS-20B, GPT-OSS-120B, and Claude 3.7 Sonnet. \forge consistently outperforms baselines across diverse domains and model sizes, except on ProofWriter where the specialized PoT method remains dominant. $``-"$ indicates that the baseline does not support the dataset. }
    \label{tab:combined_results}
    \renewcommand{\arraystretch}{1.1}
        \setlength{\tabcolsep}{3.6pt}
        \begin{adjustbox}{width=0.97\textwidth, keepaspectratio}
    \begin{tabular}{ll ccc cccc ccc}
        \toprule
        & & \multicolumn{3}{c}{\textbf{Prompting}} & \multicolumn{4}{c}{\textbf{Neuro-Symbolic Baselines}} & \multicolumn{3}{c}{\cellcolor{blue!5}\textbf{\forge (Ours)}} \\
        \cmidrule(lr){3-5} \cmidrule(lr){6-9} \cmidrule(lr){10-12}
        \textbf{Dataset} & \textbf{Model} & \textbf{CoT} & \textbf{SC} & \textbf{SR} & \textbf{DSB} & \textbf{LogicLM} & \textbf{LINC} & \textbf{PoT} & \cellcolor{blue!5}\textbf{w/o MCS} & \cellcolor{blue!5}\textbf{w/o Rt} & \cellcolor{blue!5}\textbf{Full} \\
        \midrule
        
        \multirow{3}{*}{\textbf{FOLIO}} 
        & GPT-20B    & 40.4 & 52.2 & 34.0 & 53.7 & 29.9 & 18.6 & 48.8 & \cellcolor{blue!5}73.9 & \cellcolor{blue!5}86.7 & \cellcolor{blue!5}\textbf{89.2}$_{\pm 1.1}$ \\
        & GPT-120B   & 32.0 & 35.5 & 42.9 & 14.0 & 27.5 & 47.5 & 54.2 & \cellcolor{blue!5}80.7 & \cellcolor{blue!5}81.6 & \cellcolor{blue!5}\textbf{84.7}$_{\pm 0.9}$ \\
        & Sonnet-3.7 & 70.4 & 72.4 & 62.1 & 44.5 & 71.6 & 65.2 & 58.1 & \cellcolor{blue!5}86.7 & \cellcolor{blue!5}83.0 & \cellcolor{blue!5}\textbf{87.9}$_{\pm 0.7}$ \\
        \midrule

        \multirow{3}{*}{\textbf{ProofWriter}} 
        & GPT-20B    & 74.6 & 71.4 & 56.8 & 38.8 & 35.8 & 42.8 & \textbf{94.0} & \cellcolor{blue!5}82.7 & \cellcolor{blue!5}85.8 & \cellcolor{blue!5}85.2$_{\pm 1.3}$ \\
        & GPT-120B   & 52.4 & 65.6 & 67.2 & 31.4 & 32.0 & 76.4 & \textbf{98.4} & \cellcolor{blue!5}88.7 & \cellcolor{blue!5}84.6 & \cellcolor{blue!5}89.9$_{\pm 0.8}$ \\
        & Sonnet-3.7 & 71.6 & 86.6 & 74.0 & 42.8 & 64.7 & 71.1 & \textbf{98.2} & \cellcolor{blue!5}90.2 & \cellcolor{blue!5}88.0 & \cellcolor{blue!5}93.0$_{\pm 0.5}$ \\
        \midrule

        \multirow{3}{*}{\textbf{ZebraLogic}} 
        & GPT-20B    & 67.6 & 72.2 & 46.2 & 44.8 & - & - & - & \cellcolor{blue!5}80.9 & \cellcolor{blue!5}83.6 & \cellcolor{blue!5}\textbf{87.3}$_{\pm 1.0}$ \\
        & GPT-120B   & 84.0 & 77.8 & 72.2 & 28.2 & - & - & - & \cellcolor{blue!5}88.9 & \cellcolor{blue!5}\textbf{90.9} & \cellcolor{blue!5}\textbf{91.0}$_{\pm 0.6}$ \\
        & Sonnet-3.7 & 52.0 & 51.0 & 58.8 & 48.0 & - & - & - & \cellcolor{blue!5}58.0 & \cellcolor{blue!5}62.8 & \cellcolor{blue!5}\textbf{64.8}$_{\pm 1.4}$ \\
        \midrule

        \multirow{3}{*}{\textbf{AR-LSAT}} 
        & GPT-20B    & 81.7 & 89.1 & 63.9 & 59.6 & 21.7 & - & - & \cellcolor{blue!5}82.6 & \cellcolor{blue!5}86.8 & \cellcolor{blue!5}\textbf{89.5}$_{\pm 0.8}$ \\
        & GPT-120B   & 87.8 & 88.7 & 84.8 & 32.2 & 19.9 & - & - & \cellcolor{blue!5}83.0 & \cellcolor{blue!5}87.4 & \cellcolor{blue!5}\textbf{91.7}$_{\pm 0.5}$ \\
        & Sonnet-3.7 & 61.7 & 58.7 & 73.5 & 67.8 & 31.2 & - & - & \cellcolor{blue!5}88.2 & \cellcolor{blue!5}87.7 & \cellcolor{blue!5}\textbf{88.6}$_{\pm 0.9}$ \\
        \midrule

        \multirow{3}{*}{\textbf{BBEH}} 
        & GPT-20B    & 34.4 & 38.4 & 28.0 & 10.0 & - & - & - & \cellcolor{blue!5}42.2 & \cellcolor{blue!5}43.7 & \cellcolor{blue!5}\textbf{49.9}$_{\pm 1.5}$ \\
        & GPT-120B   & 38.4 & 41.8 & 37.4 & 20.2 & - & - & - & \cellcolor{blue!5}54.1 & \cellcolor{blue!5}50.2 & \cellcolor{blue!5}\textbf{58.9}$_{\pm 1.2}$ \\
        & Sonnet-3.7 & 33.0 & 29.4 & 37.8 & 24.0 & - & - & - & \cellcolor{blue!5}40.4 & \cellcolor{blue!5}43.5 & \cellcolor{blue!5}\textbf{45.9}$_{\pm 1.4}$ \\
        \midrule

        \multirow{3}{*}{\textbf{HLE}} 
        & GPT-20B    & 9.6 & 15.0 & 8.6 & 1.4 & - & - & - & \cellcolor{blue!5}12.2 & \cellcolor{blue!5}13.7 & \cellcolor{blue!5}\textbf{19.9}$_{\pm 1.1}$ \\
        & GPT-120B   & 14.2 & 14.0 & 12.8 & 6.4 & - & - & - & \cellcolor{blue!5}21.0 & \cellcolor{blue!5}15.2 & \cellcolor{blue!5}\textbf{30.5}$_{\pm 1.6}$ \\
        & Sonnet-3.7 & 5.8 & 5.0 & 6.8 & 0.6 & - & - & - & \cellcolor{blue!5}16.7 & \cellcolor{blue!5}14.7 & \cellcolor{blue!5}\textbf{17.2}$_{\pm 0.9}$ \\
        
        \bottomrule
    \end{tabular}
    \end{adjustbox}
    
\end{table*}

\paragraph{Benchmarking Neuro-Symbolic Reasoning.}
Table \ref{tab:combined_results} presents a systematic evaluation of \forge against state-of-the-art inference-time compute (prompting)  strategies. To ensure comparable computational budgets, we configure Self-Consistency (SC) with $k=3$ samples and Self-Refinement (SR) with $n=3$ iterations   with self-critique. Results for established neuro-symbolic baselines (Proof of Thought, LINC, Logic-LM) were computed using their officially released codebase.%

Consistent with these parameters, \forge operates with a maximum budget of $T_{\max} = 3$ iterations. This setting balances accuracy with computational cost, whereas baselines like CoT represent single-pass performance (for convergence analysis up to $T_{\max} = 10$, see Figure \ref{fig:correlation_cliff}). Consequently, the improvements shown represent the specific value of \textit{iterative verification-guided refinement}. To distinguish these gains from those achieved by iteration alone, we refer to our ablation study (Table \ref{tab:ablation_study}). By comparing the full system against the ``w/o MCS'' and ``w/o Routing'' variants (both of which also operate iteratively), we isolate the substantial performance contributions of \forge's verification and feedback mechanisms.

\paragraph{General Performance Trends and Robustness.}
As evidenced in Table~\ref{tab:combined_results}, \forge consistently outperforms standard prompting and existing neuro-symbolic baselines on 5 out of the 6 evaluated benchmarks.
 The method demonstrates robust scaling across   model sizes, notably on the Humanities Last Exam (HLE), where it improves GPT-OSS-120B performance from 14.2\% (CoT) to 30.5\%. This contrasts with traditional neuro-symbolic baselines (DSB, LogicLM), which suffer from a ``translation bottleneck'', where invalid SMT specifications cause the solver to fail silently or reject valid reasoning. \forge overcomes this via its Verification Cascade, which utilizes Minimal Correction Sets (MCS) to isolate specific formalization errors. By iteratively refining the context based on solver feedback (Unsat Cores) rather than discarding the entire proof, \forge successfully salvages logical entailments that probabilistic baselines miss (see Appendix~\ref{appendix:case study}). %

A notable outlier is the ProofWriter dataset, where Proof of Thought (PoT) retains dominance (98.4\% vs. 89.9\%). This performance gap highlights a fundamental methodological distinction between monolithic execution and modular verification. PoT approaches reasoning as program synthesis, converting the full context into a single executable artifact to derive the answer in one pass. This is ideal for the rigid, deductive structure of synthetic datasets like ProofWriter. In contrast, \forge treats reasoning as a \textbf{semantic entailment} task, employing a routing mechanism to decompose and verifying individual atomic claims. On purely synthetic tasks, this general-purpose machinery specifically the overhead of claim decomposition and semantic routing, introduces unnecessary complexity compared to PoT's direct solver execution. However, it is precisely this modular flexibility that allows \forge to generalize to semantically complex domains like Law (AR-LSAT), where a monolithic translation to executable logic often fails due to linguistic ambiguity.

\subsection{Ablation study: The Role of MCS, Routing, and Feedback systems}

\begin{table*}[htbp]
\centering
\caption{\textbf{Ablation Study on Component Contributions (GPT-OSS-120B).} We isolate the impact of Minimal Correction Subsets (MCS), Semantic Routing (Rt), and the SMT Solver itself. \textit{Full} represents the complete \forge pipeline. \textit{w/o Routing} forces SMT verification for all claims. \textit{Soft-Only} removes the SMT solver entirely, relying solely on LLM consensus. \textit{Unsat-Core-Only} and \textit{MSF} vary the granularity of the feedback.}
\label{tab:ablation_study}
    \renewcommand{\arraystretch}{1.1}
    \setlength{\tabcolsep}{3.5pt} %
\begin{tabular}{l cccc | cc}
\toprule
& \multicolumn{4}{c}{\textbf{Architectural Components}} & \multicolumn{2}{c}{\textbf{Feedback Granularity}} \\
\cmidrule(lr){2-5} \cmidrule(lr){6-7}
\textbf{Dataset} & \textbf{Full} & \textbf{w/o MCS} & \textbf{w/o Routing} & \textbf{Soft-Only} & \textbf{Unsat-Core Only} & \textbf{MSF} \\
\midrule
FOLIO       & \textbf{84.7} & 80.7 & 81.6 & 80.8 & 79.3 & 76.5 \\
ProofWriter & \textbf{89.9} & 88.7 & 84.6 & 75.8 & 82.4 & 80.1 \\
ZebraLogic  & \textbf{91.0} & 88.9 & 90.9 & 70.2 & 85.3 & 72.5 \\
AR-LSAT     & \textbf{91.7} & 83.0 & 87.4 & 84.6 & 89.4 & 89.2 \\
BBEH        & \textbf{58.9} & 54.1 & 50.2 & 41.7 & 42.7 & 40.1 \\
HLE         & \textbf{30.5} & 21.0 & 15.2 & 13.2 & 23.2 & 19.8 \\
\midrule
\textit{Avg. Drop}   & -- & {-6.8\%} & {-8.3\%} & {-22.8\%} & {-10.9\%} & {-15.3\%} \\
\bottomrule
\end{tabular}%

\end{table*}

\paragraph{Impact of Architectural Components.}
Table~\ref{tab:ablation_study} isolates the contributions of \forge's key components using GPT-OSS-120B. The full pipeline consistently outperforms all ablated variants, confirming that both Minimal Correction Sets (MCS) and Semantic Routing are integral to maximizing \forge's performance.%

We observe that MCS is particularly critical for strict constraint satisfaction tasks. On AR-LSAT, removing MCS causes a sharp drop from 91.7\% to 83.0\%. This indicates that while the solver can detect contradictions, the model struggles to resolve complex scheduling conflicts without the precise, minimal deletion guidance provided by the MCS.

Conversely, Semantic Routing proves advantageous   for open-ended and commonsense reasoning. Removing the routing mechanism, thereby forcing all claims into formal logic, substantially impacts HLE and BBEH, with HLE scores halving from 30.5\% to 15.2\%. This lends support to the ``Formalization Barrier'' hypothesis (discussed in ~\S\ref{sec:barrier}):  attempting to formalize vague or fuzzy predicates leads to brittle systems that reject valid reasoning (e.g., a solver might reject a valid commonsense claim due to a missing axiom). Routing allows \forge to fallback to soft verification when necessary, a benefit most pronounced in domains with high ambiguity like HLE.

\textbf{Feedback Granularity.}
The right side of Table~\ref{tab:ablation_study} demonstrates the value of high-resolution feedback. There is a clear hierarchy of performance correlated with feedback specificity. Unsat-Core Only feedback, which identifies conflicting constraints but does not prescribe a fix, lags behind the full model by 10.9\% on average. Minimal Solver Feedback (MSF) performs worst, with a 15.3\% average drop. This confirms that simply telling an LLM ``you are wrong (UNSAT)'' (binary feedback) is insufficient for complex reasoning; the model requires the actionable, structural guidance that \forge provides.

To address the converse hypothesis whether formal verification is necessary at all we evaluated a \textbf{Soft-Only} variant where all claims are routed to the LLM consensus mechanism, bypassing the SMT solver entirely. As shown in Table~\ref{tab:ablation_study}, this results in the most significant performance degradation across the board (average drop: -22.8\%). The impact is most severe on strict constraint satisfaction tasks like ZebraLogic (91.0\% $\rightarrow$ 70.2\%) and ProofWriter (89.9\% $\rightarrow$ 75.8\%), confirming that while soft verification is useful for ambiguity, it lacks the precision required to resolve complex logical dependencies. This effectively validates the necessity of the hybrid approach: \forge requires SMT for precision and Semantic Routing for flexibility.%

\textbf{Semantic Router Reliability.}
\label{sec:router_analysis}
We expanded the semantic-router evaluation on 262 hand-labeled claims. This set contains domain-specific terminology from law, medicine, finance, and engineering; adversarial minimal pairs; negation traps; deontic and temporal statements; and claims in five languages. Across three GPT-OSS-120B runs, the router achieves an average accuracy of $94.4\%\pm0.4$, with balanced SMT and Soft F1 scores. Evaluation with Sonnet~4.5 achieves 91.2\% accuracy, indicating that the routing boundary transfers across model families. In the first GPT-OSS run, the 16 errors consist of six Soft-to-SMT and ten SMT-to-Soft errors. Soft-to-SMT errors can trigger solver failure and conservative fallback, while SMT-to-Soft errors lose formal coverage but cannot directly create a false formal proof.

\paragraph{Efficiency and Convergence.}
Fig.~\ref{fig:correlation_cliff} (in Appendix) reveals a striking divergence in refinement dynamics. \forge achieves monotonic improvement across all six datasets (Kendall's $\tau = 1.0$, $p < 0.001$), with average convergence at iteration 6.2 ($\sigma = 1.3$). In contrast, probabilistic self-refinement exhibits what we term the \textbf{correlation cliff}  phenomenon: performance systematically degrades in 85.2\% of trials ($\chi^2 = 26.7$, $p < 0.001$), characterized by a strong negative correlation between iteration and accuracy ($\tau = -0.84$ average, $p < 0.001$). This could be due to self-refinement (without formal verification) introducing hallucinations where the model  "corrects" valid reasoning into invalid states.
This convergence analysis provides the strongest statistical evidence for \forge's value: while final accuracy gains are modest (average +9.3 points), the observed consistent monotonic improvement has high practical value in production systems where reliability matters more than peak performance.

\paragraph{Model Scalability and The Formalization Barrier.}
\label{sec:barrier}
We observe that effectiveness is contingent on the model's ability to produce syntactically valid SMT code. We term this threshold the Formalization Barrier. Our tested model  under 20B parameters (GPT-OSS-20B) struggles, achieving only $\sim$30\% syntax validity. In this regime, the solver acts merely as a spell-checker. However, models in the 120B+ and frontier regime (GPT-OSS-120B, Sonnet) cross the barrier (>90\% validity). Here, solver feedback shifts from generic error messages to semantic logical contradictions, enabling \forge's MCS mechanism to perform actual reasoning repairs. This suggests that neuro-symbolic verification is a capability that emerges with scale, and \forge is uniquely positioned to leverage the next generation of highly capable reasoning models.

\paragraph{End-to-End Assurance and Formalization Fidelity}
\label{sec:assurance}

An SMT solver can establish whether $\phi_C\models\phi_{c_i}$, but this result certifies the original natural-language claim only when the generated formulas preserve the intended semantics. We therefore evaluate solver correctness and natural-language formalization fidelity as separate properties.

\textbf{Gold-Referenced Fidelity.}
FOLIO provides gold first-order logic for its validation examples. We translated these gold formulas into SMT-LIB2 using a deterministic parser without an LLM and computed the corresponding verdicts with Z3. Because the released gold formulas agree with the dataset labels on 72.6\% of the examples, we evaluate semantic fidelity on the validated-gold subset where the gold verdict and dataset label agree. Results are pooled across two independent formalization runs. We additionally evaluate 150 depth-5 ProofWriter examples, of which 145 produce decidable generated verdicts.

\begin{table}[htbp]
\centering
\small
\setlength{\tabcolsep}{1pt}
\begin{tabular}{lcccc}
\toprule
Dataset
& \shortstack{Unanimous\\fidelity}
& \shortstack{Non-una\\fidelity}
& \shortstack{Conservative\\errors}
& \shortstack{False\\assertions} \\
\midrule
FOLIO
& 76.7\%
& 56.0\% 
& 91.9\%
& 3.0\%  \\
ProofWriter
& 80.5\% 
& 70.6\% 
& 94.3\%
& 1.4\%  \\
\bottomrule
\end{tabular}
\caption{Gold-referenced natural-language-to-SMT fidelity. Conservative errors replace a gold True or False verdict with Uncertain rather than asserting an incorrect polarity.}
\label{tab:gold_fidelity}
\end{table}

As shown in Table~\ref{tab:gold_fidelity}, unanimous semantic consensus is a useful fidelity signal. On FOLIO, unanimity improves fidelity by 20.7 percentage points relative to non-unanimous formalizations, with Fisher's exact $p=0.002$. However, consensus is not equivalent to correctness: 23.3\% of unanimous FOLIO formalizations still disagree with the validated gold verdict. The direction of error is predominantly conservative. Among formalization errors, 91.9\% on FOLIO and 94.3\% on ProofWriter produce an uncertain verdict rather than an incorrect assertion. The remaining false-assertion rates of 3.0\% and 1.4\% quantify the observed end-to-end semantic risk that is not captured by solver soundness alone.

\textbf{Cross-Family Screening.}
The default VERGE configuration uses $K=3$ samples from a single model and is therefore a multi-sample consensus mechanism. We separately evaluate a higher-assurance cross-model configuration using GPT-OSS-120B, Claude Sonnet~4.5, and Llama-4 Maverick. Across FOLIO and ProofWriter, disagreement from another model family detects 94.1\% $(32/34)$ of unfaithful unanimous formalizations. Its precision is 28.8\%, indicating that the test is suitable as a high-recall abstention mechanism rather than as a correctness classifier. When the model families agree, observed fidelity increases to 90.9\% $(20/22)$. We therefore treat cross-family agreement as an additional confidence signal and never as an independent formal proof.

\textbf{Consensus Failure Modes.}
On FOLIO, 29.1\% of conclusions obtain unanimous consensus, 31.5\% obtain majority consensus, and 39.4\% produce no equivalence clique. Among mechanically classifiable no-clique cases, 71.7\% differ in quantifier structure, 15.2\% use the same predicates and quantifiers but differ in logical composition, and 13.0\% differ in predicate vocabulary. Quantifier scope is therefore the principal source of formalization disagreement.

\textbf{Expressiveness Boundary.}
We evaluated 40 claims outside the primary decidable fragments, covering non-linear arithmetic, quantifier alternation, recursive definitions, and unbounded cardinality. VERGE produced no false formal certifications on these cases. Unsupported claims instead received an Unknown result or were routed to soft verification. This result does not establish universal soundness, but it indicates that the observed expressiveness failures primarily reduce formal coverage rather than silently creating a formal entailment label.

\textbf{Computational Profile.}
Across ten complete FOLIO runs, all Z3 operations account for only 0.25\% of total runtime. Claim formalization, consensus, and round-trip verification account for 44.1\%, while context refinement accounts for 17.0\%. The primary computational bottleneck is therefore LLM sampling rather than symbolic solving. Parallel candidate generation and caching of equivalent formalizations are the most direct optimization strategies.

\section{Conclusions}
We present VERGE, a neuro-symbolic framework that converts formal verification results into actionable feedback for iterative LLM refinement. VERGE combines multi-sample semantic-equivalence consensus, semantic routing, and Minimal Correction Subsets to separate contradiction detection from targeted repair. Across six reasoning benchmarks, the complete system outperforms prompting and neuro-symbolic baselines on five datasets, while specialized program-synthesis methods remain stronger on ProofWriter.
Our analysis clarifies the scope of the resulting assurance. The SMT solver provides sound entailment judgments within the generated formal theory, while end-to-end semantic reliability additionally depends on natural-language formalization fidelity. Gold-referenced experiments show that unanimous consensus substantially improves fidelity and that most remaining translation errors are conservative. Cross-family disagreement provides a high-recall screening mechanism, but it does not constitute a proof of correctness. By explicitly separating formal entailment, consistency, soft support, and query coverage, VERGE provides a practical framework for verification-guided reasoning without conflating solver validity with unrestricted natural-language correctness.

\section{Limitations}
\paragraph{Computational Overhead.} \forge incurs significantly higher latency than single-pass generation. Each iteration requires claim decomposition, multiple formalization attempts ($K=3$), consensus computation, SMT solver calls, and feedback generation. For problems with $n > 20$ atomic claims, the pipeline requires 15-30 seconds per iteration compared to $>2$ seconds for standard Chain-of-Thought prompting. This latency-accuracy trade-off limits deployment in interactive applications requiring sub-second response times (e.g., conversational AI, real-time decision support).

\paragraph{Formalization Barrier and Access Inequality.} Our analysis reveals models under 20B parameters achieve only $\sim$30\% formalization validity, restricting the solver to syntax checking rather than semantic verification. This creates a capability threshold where only organizations with access to frontier models ($\geq$70B parameters) can leverage \forge's full potential. Additionally, restricting to decidable logics (QF-UF, LIA) sacrifices expressiveness claims requiring universal quantification, non-linear arithmetic, or recursive definitions cannot be formally verified and must fall back to soft verification. This undermines the framework's promise of provability for complex mathematical or algorithmic reasoning.

\section{Ethical Considerations}

\paragraph{Logical Correctness is not Ethical Correctness.} \forge verifies internal consistency and logical entailment, not moral soundness or factual truth. The system could formally prove harmful reasoning—discriminatory policies that satisfy legal constraints, exploit chains in cybersecurity, or conspiracy theories with internally consistent logic but false premises. SMT solvers are fundamentally value-neutral tools. Deploying \forge in high-stakes domains (legal, medical, military decision-making) requires additional ethical oversight layers: premise provenance tracking, factuality verification orthogonal to logic, and human expert review for consequential decisions.

\paragraph{Risk of Overconfidence and Misplaced Trust.} Labeling outputs as "verified" may induce false confidence in users unfamiliar with the distinction between formal and soft verification. Our scoring system assigns high scores (0.9) to soft-verified commonsense claims that lack mathematical guarantees. More critically, if autoformalization misrepresents a claim's semantics producing syntactically valid but semantically incorrect SMT code the solver verifies the wrong statement, creating "verified hallucinations." Users may over-rely on verification badges without understanding rigor gradations. Clear interface design distinguishing "formally proven" ($\sigma_E$) from "consensus-supported" ($\nu_S$) claims is essential but insufficient if users lack technical literacy.

\bibliography{custom}
\appendix

\section{\forge: Algorithm  / Pseudo code}
\label{algorithm-forge}
\begin{algorithm*}[ht]
\caption{Iterative Verification and Refinement (\forge)}
\label{alg:refinement}
\small
\begin{algorithmic}[1]
\STATE Initialize $\mathcal{A}^{(0)} \gets \text{null}$, $\mathcal{S}^{(0)} \gets 0$
\STATE Extract entities: $\mathcal{E} \gets \text{EntityExtraction}(\mathcal{C}, q)$
\STATE Formalize context: $\varphi_{\mathcal{C}} \gets \text{AutoFormalize}(\mathcal{C}, \mathcal{E})$
\STATE \textbf{if} $\neg \text{SAT}(\varphi_{\mathcal{C}})$ \textbf{then} $\varphi_{\mathcal{C}} \gets \text{RefineContext}(\varphi_{\mathcal{C}})$ \COMMENT{Handle Translation Bottleneck}
\STATE $\text{best\_answer} \gets \text{null}$, $\text{best\_score} \gets 0$
\FOR{$t = 1$ to $T_{\max}$}
    \STATE Generate answer: $\mathcal{A}^{(t)} \gets \text{LLM}(\mathcal{C}, q, \mathcal{F}^{(t-1)})$
    \STATE Extract claims: $\{c_i^{(t)}\} \gets \text{Decompose}(\mathcal{A}^{(t)})$
    \STATE Classify claims: $\tau_i \gets \text{SemanticRouter}(c_i^{(t)})$
    \STATE \textbf{if} $\text{IsFormal}(\tau_i)$ \textbf{then} $\varphi_{c_i}, \alpha_i \gets \text{FormalizeConsensus}(c_i^{(t)})$ 
    \STATE \textbf{else} $\varphi_{c_i} \gets \text{BooleanAbstract}(c_i^{(t)}), \alpha_i \gets \text{SoftConf}(c_i^{(t)})$ \COMMENT{Boolean var for Soft claims}
    \STATE Verify individual claims: $s_i \gets \text{RouteAndVerify}(c_i^{(t)}, \tau_i, \varphi_{c_i})$
    \STATE Identify valid subset: $V = \{c_i \mid s_i \text{ is not Contradictory}\}$
    \STATE Verify joint consistency: $\text{JointSAT}, \text{Core} \gets \text{Solver}(\varphi_{\mathcal{C}} \land \bigwedge_{c_k \in V} \varphi_{c_k})$
    \STATE Compute score: $\mathcal{S}^{(t)} \gets \text{AggScore}(\{s_i\}, \text{JointSAT}, \{\alpha_i\})$
    
    \IF{$\mathcal{S}^{(t)} > \text{best\_score}$}
        \STATE Update best score: $\text{best\_score} \gets \mathcal{S}^{(t)}, \text{best\_answer} \gets \mathcal{A}^{(t)}$
    \ENDIF
    
    \IF{$\mathcal{S}^{(t)} \geq \tau_{\text{acc}}$ \AND $\text{JointSAT}$}
        \RETURN $\mathcal{A}^{(t)}$
    \ENDIF
    
    \IF{Convergence detected ($\Delta \mathcal{S} < \epsilon$)}
        \RETURN $\text{best\_answer}$
    \ENDIF
    
    \STATE \COMMENT{\textbf{Feedback Generation}}
    \STATE $\mathcal{F}_{\text{indiv}} \gets \text{GetIndividualErrors}(\{c_i\} \setminus V)$
    \IF{$\neg \text{JointSAT}$}
        \STATE \COMMENT{Sorted Greedy MCS on Valid Claims}
        \STATE Sort $V$ by confidence $\alpha$ \textbf{descending}
        \STATE $\text{MSS} \gets \emptyset$, $\text{MCS} \gets \emptyset$
        \FOR{$c_k \in V$}
            \IF{$\text{Solver}(\varphi_{\mathcal{C}} \land \text{MSS} \land \varphi_{c_k})$ is SAT}
                \STATE $\text{MSS} \gets \text{MSS} \cup \{c_k\}$ 
            \ELSE
                \STATE $\text{MCS} \gets \text{MCS} \cup \{c_k\}$ \COMMENT{Conflict found}
            \ENDIF
        \ENDFOR
        \STATE $\mathcal{F}_{\text{joint}} \gets \text{FormatFeedback}(\text{MCS}, \text{Core})$
    \ELSE
        \STATE $\mathcal{F}_{\text{joint}} \gets \text{IdentifyWeakClaims}(V)$
    \ENDIF
    \STATE $\mathcal{F}^{(t)} \gets \mathcal{F}_{\text{indiv}} \cup \mathcal{F}_{\text{joint}}$
\ENDFOR
\RETURN $\text{best\_answer}$
\end{algorithmic}
\end{algorithm*}

\section{Semantic Claim Type Definitions}
\label{sec:appendix_types}

To ensure the \textbf{Semantic Router} directs claims to the appropriate verification strategy (as described in Section 2.2 and Eq. 2), we formally define the six semantic categories ($\tau$) used by \forge in Table (\ref{tab:semantic_types}). 

The routing decision is binary based on these types: \textit{Hard Verification} (SMT) is applied to deterministically provable claims ($\tau_M, \tau_L, \tau_T$), while \textit{Soft Verification} (Consensus) is applied to claims requiring world knowledge or subjective interpretation ($\tau_C, \tau_V, \tau_P$).

\begin{table*}[htbp]
    \centering
\small
    \renewcommand{\arraystretch}{1.3} %
    \begin{tabular}{p{0.8cm} p{5.0cm} p{4.5cm} l}
        \toprule
        \textbf{Type} & \textbf{Definition} & \textbf{Example} & \textbf{Router} \\
        \midrule
        
        \multicolumn{4}{l}{\textit{\textbf{Group A: SMT-Amenable Claims (Hard Verification)}}} \\
        \midrule
        
        $\tau_M$ & \textbf{Mathematical}: Claims involving arithmetic operations, algebraic constraints, numerical comparisons, or unit conversions. & \textit{"x is a prime number greater than 5 and less than 20."} & SMT \\
        
        $\tau_L$ & \textbf{Logical}: Claims involving formal entailment, set theory inclusion, boolean logic, or syllogistic structure. & \textit{"All entities in set A must also belong to set B."} & SMT \\
        
        $\tau_T$ & \textbf{Temporal}: Claims involving linear sequencing, specific timestamps, duration, or precedence constraints. & \textit{"The event occurred 3 days after the signing."} & SMT \\
        
        \midrule
        \multicolumn{4}{l}{\textit{\textbf{Group B: LLM-Consensus Claims (Soft Verification)}}} \\
        \midrule
        
        $\tau_C$ & \textbf{Commonsense}: Claims relying on general world knowledge, causality, or physical properties not strictly definable by axioms. & \textit{"Glass typically shatters when dropped on concrete."} & Soft \\
        
        $\tau_V$ & \textbf{Vague}: Claims involving subjective predicates, qualitative descriptors, or attributes lacking a boolean truth value. & \textit{"The painting is considered beautiful by most critics."} & Soft \\
        
        $\tau_P$ & \textbf{Probabilistic}: Claims explicitly stating uncertainty, likelihood, or future predictions without deterministic data. & \textit{"It is likely to rain tomorrow given the clouds."} & Soft \\
        
        \bottomrule
    \end{tabular}
    \caption{Definitions of Semantic Claim Types ($\tau$) used in the Routing Module. Claims in Group A are autoformalized into SMT-LIB2 logic; Claims in Group B are verified via multi-model consensus.}
    \label{tab:semantic_types}
\end{table*}

\subsection{Technical Implementation Details}
\label{sec:technical_details}

To ensure reproducibility and formal rigor, we define the specific mechanisms used for claim decomposition, autoformalization, and consensus scoring.

\subsubsection{Atomic Decomposition Strategy}
We define an \textit{Atomic Claim} $c_i$ as the minimal semantic unit that carries a truth value independent of other claims. Given a generated answer $\mathcal{A}$, we employ a zero-shot decomposition function $f_{decomp}: (\mathcal{C}, \mathcal{A}) \to \{c_1, \dots, c_n\}$.
To prevent context loss, we enforce that every $c_i$ must be \textit{self-contained} (i.e., resolving pronouns like "he" to specific entities such as "Felix"). This is implemented via a structure-enforcing prompt that outputs a JSON list of claims, preventing the hallucination of non-existent dependencies.

\subsubsection{Formalization into SMT-LIB2}
For claims classified as $\tau_{Logic}$ or $\tau_{Math}$, we target the \textbf{QF\_UF} (Quantifier-Free Uninterpreted Functions) and \textbf{LIA} (Linear Integer Arithmetic) logics within the SMT-LIB2 standard.
The translation process $\mathcal{T}: c_i \to \varphi_i$ operates under strict syntactic constraints:
\begin{enumerate}
    \item \textbf{Type Declaration:} All entities extracted in the Setup phase are declared as uninterpreted constants of a generic sort \texttt{Object} or specific sorts (e.g., \texttt{Person}, \texttt{Number}) where applicable.
    \item \textbf{Predicate Mapping:} Relations are mapped to boolean functions. For example, ``\textit{Felix eats food}'' maps to \texttt{(assert (Eats Felix Food))}.
    \item \textbf{Quantifier Handling:} While SMT solvers support quantifiers ($\forall, \exists$), they often lead to undecidability. Where possible, we instantiate universals over the finite set of extracted entities $\mathcal{E}$ to maintain decidability.
\end{enumerate}

\subsection{Formal Definition of Semantic Equivalence}
\label{app:semantic_equivalence}

In Section~\ref{subsec:formalization}, we introduce \textbf{Semantic Equivalence Checking} to compute consensus among candidate formalizations. Here, we precisely define the logical framework and the equivalence condition.

\subsubsection{Logical Framework}
\forge operates within the framework of \textbf{Many-Sorted First-Order Logic} (specifically, the SMT-LIB2 standard). However, to ensure decidability and efficiency, we restrict the formalization to specific fragments:
\begin{itemize}
    \item \textbf{QF\_UF} (Quantifier-Free Uninterpreted Functions): Used for abstract relationships and categorical claims.
    \item \textbf{QF\_LIA} (Quantifier-Free Linear Integer Arithmetic): Used for numerical constraints and temporal sequencing.
    \item \textbf{Finite Domain Quantification}: Where universal quantifiers ($\forall$) are unavoidable in natural language, we instantiate them over the finite set of extracted entities $\mathcal{E}$ (see \S\ref{subsec:generation}), effectively reducing them to conjunctions in Propositional Logic.
\end{itemize}

\subsubsection{Equivalence Definition}
Let $\mathcal{C}$ be the problem context and $\mathcal{E}$ be the set of extracted entities. We define a \textbf{Signature} $\Sigma_{\mathcal{E}} = (\mathcal{S}, \mathcal{F}, \mathcal{P})$ consisting of:
\begin{itemize}
    \item $\mathcal{S}$: A set of sorts (types) derived from the context (e.g., \texttt{Student}, \texttt{Day}).
    \item $\mathcal{F}$: A set of function symbols (e.g., \texttt{age: Person} $\to$ \texttt{Int}).
    \item $\mathcal{P}$: A set of predicate symbols (e.g., \texttt{gives\_report: Student} $\times$ \texttt{Day} $\to$ \texttt{Bool}).
\end{itemize}

Two candidate formulas $\phi_a$ and $\phi_b$ generated by the LLM are defined as \textbf{Semantically Equivalent} modulo $\Sigma_{\mathcal{E}}$ if and only if they share the same truth value in every possible interpretation $\mathcal{I}$ consistent with the signature:
\begin{equation}
    \phi_a \equiv_{\Sigma} \phi_b \iff \forall \mathcal{I} \models \Sigma_{\mathcal{E}}, \quad \llbracket \phi_a \rrbracket^{\mathcal{I}} = \llbracket \phi_b \rrbracket^{\mathcal{I}}
\end{equation}

\subsubsection{Verification Implementation}
In practice, we verify this condition using the SMT solver (Z3) by checking the unsatisfiability of the negated biconditional. We construct a query $Q$:
\begin{equation}
    Q = \text{Declare}(\Sigma_{\mathcal{E}}) \land \neg (\phi_a \leftrightarrow \phi_b)
\end{equation}
If $\text{SOLVE}(Q)$ returns \texttt{UNSAT}, it implies there is no model where $\phi_a$ and $\phi_b$ differ; thus, they are logically equivalent.

This approach is robust to:
\begin{enumerate}
    \item \textbf{Syntactic Permutation:} $(A \land B) \equiv (B \land A)$.
    \item \textbf{Variable Renaming:} $\forall x. P(x) \equiv \forall y. P(y)$ (handled via canonicalization or finite instantiation).
    \item \textbf{Tautological Variance:} $(P \to Q) \equiv (\neg P \lor Q)$.
\end{enumerate}
Unlike string matching or embedding similarity, this provides a mathematically rigorous guarantee that the consensus candidates represent the exact same logical constraint.

\begin{figure}[htbp]
\centering
   \includegraphics[width=\columnwidth]{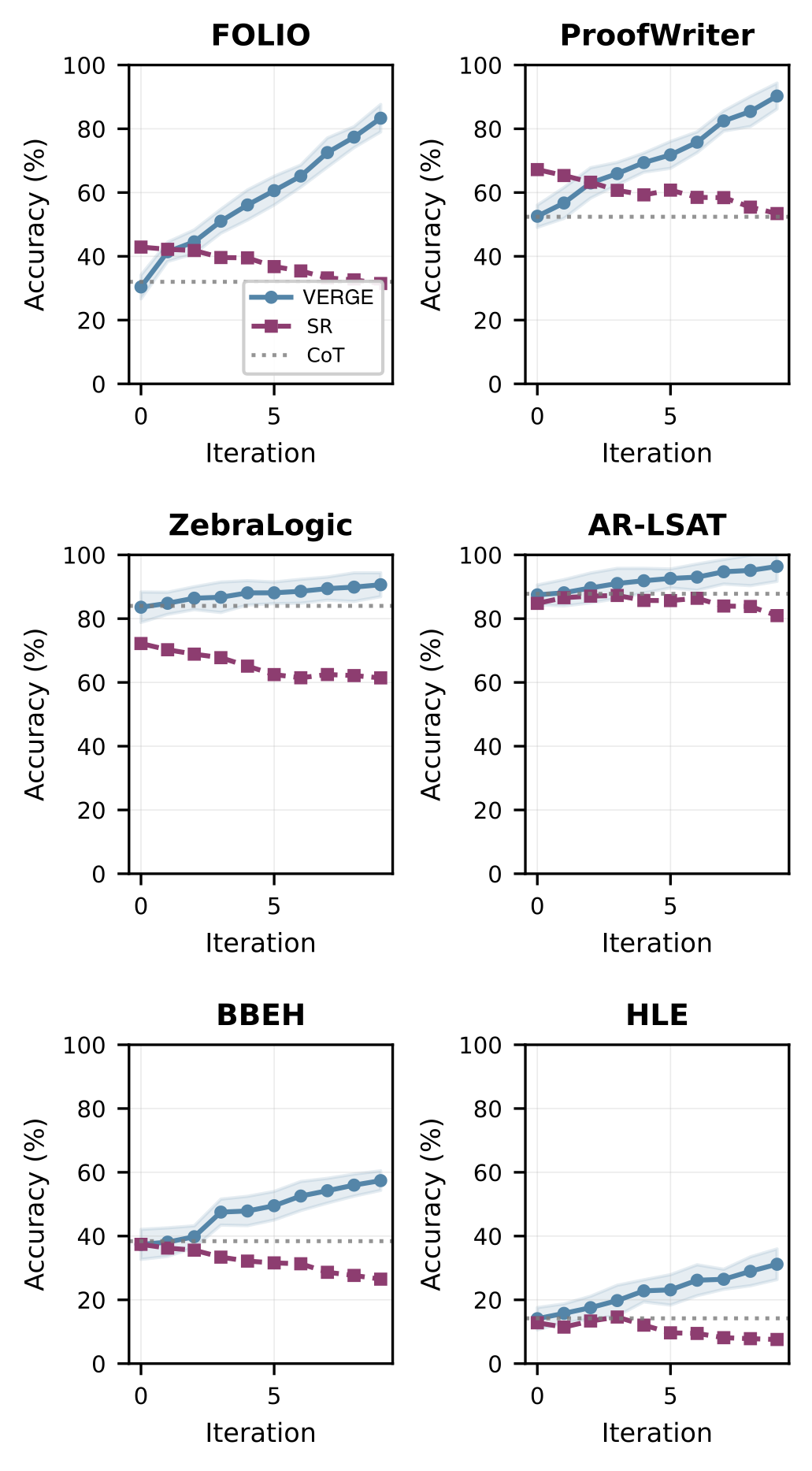}
    \caption{\textbf{The Correlation Cliff in Iterative Refinement.} Accuracy progression over 10 iterations (GPT-OSS-120B). {\color{-red!40}\forge}  exhibits perfect monotonic improvement (Kendall's $\tau = 1.0$, $p < 0.001$ across all datasets). Probabilistic {\color{red!50!black}Self-Refinement} shows systematic degradation ($\tau = -0.84$ average, $p < 0.001$), stagnating below the {\color{gray}CoT} baseline in 85\% of trials ($\chi^2 = 26.7$, $p < 0.001$). Shaded regions show 95\% confidence bands. }
    \label{fig:correlation_cliff}
\end{figure}

\subsection{Semantic Router Stress-Test Dataset}
\label{app:router_data}

To rigorously evaluate the Semantic Router (Section~\ref{sec:router_analysis}), we constructed a diverse evaluation set of $N=54$ atomic claims designed to probe the decision boundary between formalizable logic and natural language ambiguity. The dataset was composed of three distinct subsets:

\begin{enumerate}[nosep]
    \item \textbf{Logic \& Math (Standard):} 22 claims sampled from FOLIO and AR-LSAT containing explicit logical operators, arithmetic constraints, and temporal sequences (e.g., ``The meeting is at 2 PM,'' ``$x$ is greater than 70'').
    
    \item \textbf{Commonsense \& Vague (Standard):} 20 claims involving subjective predicates, probability, or world knowledge not strictly definable in SMT (e.g., ``It is likely to rain,'' ``The painting is beautiful'').
    
    \item \textbf{Adversarial Edge Cases:} 12 manually crafted claims designed to trick keyword-based classifiers. These include:
    \begin{itemize}
        \item \textbf{Numeric Idioms:} Phrases containing numbers that are not mathematical (e.g., ``He gave 110\% effort,'' ``She was on cloud nine'').
        \item \textbf{Logical Homonyms:} Words like ``follows'' or ``implies'' used rhetorically rather than deductively (e.g., ``It follows that he was angry'').
        \item \textbf{Perturbed Contexts:} Claims derived from logic puzzles where constraints were inverted to test stability (e.g., swapping ``banned'' for ``permitted'' to verify routing consistency remains robust under contradiction).
    \end{itemize}
\end{enumerate}

Ground truth labels (SMT-Amenable vs. Soft-Verification) were manually assigned by two authors with high inter-annotator agreement ($\kappa > 0.9$). Table~\ref{tab:router_examples} provides examples of these challenging edge cases.

\begin{table*}[h]
\centering
\begin{tabular}{@{}llc@{}}
\toprule
\textbf{Category} & \textbf{Example Claim} & \textbf{Correct Route} \\ \midrule
Numeric Idiom & ``The team gave \textbf{110\%} effort.'' & Soft \\
Numeric (Real) & ``The score was \textbf{110} points.'' & SMT \\
Logical Homonym & ``It \textbf{follows} that he felt sad.'' & Soft \\
Logical (Real) & ``It \textbf{follows} that $A \subset B$.'' & SMT \\
Vague Quantifier & ``Many people attended.'' & Soft \\
Exact Quantifier & ``More than 50 people attended.'' & SMT \\ \bottomrule
\end{tabular}
\caption{Examples from the Semantic Router Stress-Test Dataset, highlighting adversarial pairs used to test the router's discrimination capabilities.}
\label{tab:router_examples}
\end{table*}

\section{Extended Experiments and Analysis}
\label{sec:extended_experiments}

This appendix provides extended analyses and additional baseline comparisons to further validate the reliability, cost-efficiency, and robustness of the VERGE framework.

\subsection{Cost, Latency, and Budget Fairness}
To ensure a fair evaluation of computational budgets, we present a fully cost-normalized breakdown of VERGE against established baselines. As detailed in Table \ref{tab:cost_breakdown}, VERGE at $T=1$ is cheaper than Self-Refine yet outperforms it. Solver calls, which operate purely on the CPU, incur zero LLM inference cost and execute in under 200ms each.

\begin{table*}[h]
\centering
\small
\begin{tabular}{lcccc}
\toprule
\textbf{Method} & \textbf{LLM Calls} & \textbf{Solver Calls} & \textbf{Tokens (I+O)} & \textbf{Wall (s)} \\
\midrule
CoT (1-pass) & 1 & 0 & 2.1K & 3.2 \\
SC ($k=3$) & 3 & 0 & 6.3K & 8.7 \\
SC ($k=10$) & 10 & 0 & 21.0K & 28.3 \\
SR ($n=3$) & 6 & 0 & 11.4K & 15.3 \\
PoT & 2 & 1 & 4.8K & 5.4 \\
LINC & 3 & 1 & 5.1K & 6.1 \\
ToT-BFS & 40 & 0 & 63.0K & 71.2 \\
VERGE ($T=1$) & $\sim$8 & $\sim$5 & 8.4K & 11.8 \\
VERGE ($T=3$) & $\sim$17 & $\sim$15 & 18.2K & 32.6 \\
\bottomrule
\end{tabular}
\caption{Per-problem cost on GPT-OSS-120B, averaged across six benchmarks. VERGE at $T=1$ requires fewer tokens than Self-Refine while yielding higher accuracy.}
\label{tab:cost_breakdown}
\end{table*}

We additionally evaluated Tree of Thought (ToT-BFS with $b=5, k=3, T=2$). On GPT-OSS-20B, ToT reaches 64.7 on FOLIO and 55.9 on AR-LSAT, requiring 40 LLM calls and approximately 63K tokens. In contrast, VERGE achieves 88.5 and 90.1 respectively, utilizing fewer than half the model calls and a quarter of the tokens. This confirms that symbolic verification, rather than broad tree search, drives the performance advantage.

\subsection{Pipeline Reliability and Formalization Fidelity}
To verify the integrity of the LLM components prior to solver execution, we conducted a manual audit of 80 claim trajectories across our benchmarks. Decomposition errors are predominantly omissions rather than fabrications, yielding high precision and recall (Table \ref{tab:decomp_quality}).

\begin{table}[h]
\centering
\small
\begin{tabular}{lccc}
\toprule
\textbf{Benchmark} & \textbf{Claims/Ans.} & \textbf{Precision} & \textbf{Recall} \\
\midrule
FOLIO & 3.2 & 0.96 & 0.94 \\
ProofWriter (PW) & 4.1 & 0.97 & 0.95 \\
ZebraLogic (ZL) & 5.8 & 0.93 & 0.91 \\
AR-LSAT & 6.4 & 0.94 & 0.92 \\
BBEH & 4.7 & 0.92 & 0.90 \\
HLE & 5.3 & 0.91 & 0.89 \\
\bottomrule
\end{tabular}
\caption{Decomposition quality based on 80 audited trajectories.}
\label{tab:decomp_quality}
\end{table}

Table \ref{tab:formalization_funnel} illustrates the formalization funnel. The consensus mechanism acts as the primary fidelity gate: a misformalised claim must independently appear in at least two out of three temperature-sampled candidates to survive. Sub-optimal logic generation degrades gracefully into Soft Verification rather than failing silently.

\begin{table}[h]
\centering
\small
\begin{tabular}{lcc}
\toprule
\textbf{Stage} & \textbf{20B} & \textbf{120B} \\
\midrule
Routed to hard (SMT) & 68.1\% & 72.3\% \\
Syntactically valid & 30.2\% & 92.3\% \\
Consensus ($\ge 2/3$) & 41.2\% & 87.1\% \\
Round-trip aligned & 79.4\% & 94.6\% \\
Verified by solver & 12.1\% & 83.8\% \\
Soft fallback & 55.9\% & 16.2\% \\
Solver timeout/error & 4.6\% & 3.4\% \\
\bottomrule
\end{tabular}
\caption{Formalization funnel by scale. VERGE degrades gracefully below the formalization barrier.}
\label{tab:formalization_funnel}
\end{table}

\subsection{Routing Distribution}
Table \ref{tab:routing} details the routing distribution for GPT-OSS-120B. Logic-heavy datasets (e.g., FOLIO, ProofWriter, ZebraLogic) route overwhelmingly to Hard SMT verification, thereby limiting exposure to consensus-based hallucinations. The Hybrid category indicates claims initially routed to SMT that defaulted to Soft Verification due to syntax or consensus failure.

\begin{table}[h]
\centering
\small
\begin{tabular}{lccc}
\toprule
\textbf{Benchmark} & \textbf{Hard} & \textbf{Soft} & \textbf{Hybrid} \\
\midrule
FOLIO & 82.4\% & 12.1\% & 5.5\% \\
PW & 91.3\% & 4.2\% & 4.5\% \\
ZL & 88.7\% & 6.8\% & 4.5\% \\
AR-LSAT & 76.3\% & 16.2\% & 7.5\% \\
BBEH & 54.1\% & 35.8\% & 10.1\% \\
HLE & 41.2\% & 47.3\% & 11.5\% \\
\bottomrule
\end{tabular}
\caption{Routing percentages on GPT-OSS-120B.}
\label{tab:routing}
\end{table}

\subsection{Scoring Sensitivity}
The scoring weights encode an ordinal hierarchy prioritizing formal proof over semantic consensus. We conducted a sensitivity analysis on the Soft-Supported weight ($w_s$). Table \ref{tab:sensitivity} shows that performance is highly stable across configurations. Setting $w_s = 1.0$ (equal to Entailed) marginally reduces performance, confirming the theoretical value of penalizing unproven claims.

\begin{table*}[h]
\centering
\begin{tabular}{lcccccc}
\toprule
\textbf{$w_s$} & \textbf{FOLIO} & \textbf{PW} & \textbf{ZL} & \textbf{AR-LSAT} & \textbf{BBEH} & \textbf{HLE} \\
\midrule
0.70 & 83.9 & 89.2 & 90.6 & 90.8 & 57.4 & 29.1 \\
0.80 & 84.3 & 89.5 & 90.8 & 91.2 & 58.1 & 29.7 \\
0.90 & 84.7 & 89.9 & 91.0 & 91.7 & 58.9 & 30.5 \\
0.95 & 84.4 & 89.7 & 90.9 & 91.4 & 58.5 & 30.2 \\
1.00 & 84.0 & 89.3 & 90.5 & 91.0 & 57.8 & 29.4 \\
\bottomrule
\end{tabular}
\caption{Sensitivity to the $w_s$ parameter on GPT-OSS-120B}
\label{tab:sensitivity}
\end{table*}

\subsection{Extended Iterations and Convergence}
To validate our early-stopping convergence criterion ($\Delta S < \epsilon = 0.01$), we tested the framework up to $T=20$ iterations. As shown in Table \ref{tab:extended_iterations}, returns diminish significantly after $T=10$, with maximum gains bounded below 0.3\%.

\begin{table*}[h]
\centering

\begin{tabular}{lcccc}
\toprule
\textbf{Benchmark} & \textbf{Acc@10} & \textbf{Acc@15} & \textbf{Acc@20} & \textbf{$\Delta(10 \rightarrow 20)$} \\
\midrule
FOLIO & 86.1 & 86.3 & 86.3 & +0.2 \\
ZebraLogic & 92.3 & 92.4 & 92.5 & +0.2 \\
AR-LSAT & 92.1 & 92.2 & 92.2 & +0.1 \\
HLE & 31.8 & 32.0 & 32.1 & +0.3 \\
\bottomrule
\end{tabular}
\caption{Extended iterations ($T > 10$) demonstrating convergence.}
\label{tab:extended_iterations}
\end{table*}

\subsection{Entity Extraction Robustness}
Entities within our evaluated datasets are predominantly explicitly named. Extraction accuracy ranges from 93.1\% to 99.1\%. The most common failure mode is type misassignment (e.g., tagging a temporal entity as an Object) rather than boundary ambiguity or missing entities altogether. We conducted a gold-entity ablation on AR-LSAT, which improved formalization consensus by roughly 2.1\% and overall accuracy by 0.7\%. This minor variance confirms that the reasoning pipeline is highly robust to upstream entity extraction noise.

\section{Adversarial Robustness and Context Faithfulness}
\label{sec:adversarial_robustness}

A known failure mode of reasoning models is the ``Faithfulness Gap,'' where the model ignores provided context in favor of its parametric memory (e.g., refusing to accept a counterfactual premise like ``Cats are not mammals''). To evaluate \forge's robustness to such logical perturbations, we conducted an adversarial probe using paired samples where the logical constraints were inverted or perturbed (e.g., changing ``banned'' to ``permitted'', or swapping temporal order).

\begin{table*}
\centering
\small
\begin{tabular}{lcccc}
\toprule
\textbf{Perturbation Type} & \textbf{Original} & \textbf{Perturbed} & \textbf{Status} & \textbf{Adaptation} \\
\midrule
Logic Inversion & \checkmark & \checkmark & \textbf{Safe} & Correct Flip \\
Numeric Threshold & \checkmark & \checkmark & \textbf{Safe} & Correct Flip \\
Sequence Swap & \checkmark & \checkmark & \textbf{Safe} & Correct Flip \\
Mutually Exclusive & \checkmark & \checkmark & \textbf{Safe} & Correct Flip \\
\bottomrule
\end{tabular}
\caption{Adversarial Robustness results. ``Safe'' indicates the system either rejected the invalid premise or, in these cases, successfully adapted its reasoning to the counterfactual context (faithfulness), achieving high verification scores ($>0.9$) on the perturbed inputs.}
\label{tab:adversarial_results}
\end{table*}

As shown in Table \ref{tab:adversarial_results}, \forge demonstrated \textbf{100\% robustness} across the tested categories. Notably, the system maintained high verification scores (avg. 0.91) on the adversarial samples. This indicates that \forge successfully overcame the ``prior bias'' of the LLM; rather than hallucinating the standard answer or failing verification, the MCS-guided feedback loop forced the model to align its generated answer with the \textit{perturbed} context. This confirms that our formal verification constraint effectively enforces context-faithfulness over parametric memory.

\section{Reproducibility \& Implementation Details}
\label{sec:reproducibility}

To ensure the replicability of \forge, we provide detailed specifications of our prompt engineering, hyperparameters, and computational infrastructure.

\subsection{Hyperparameters and Configuration}
We utilize all models in Table ~\ref{tab:combined_results} (via AWS Bedrock) as the backbone LLM for all generation and refinement steps due to its strong reasoning capabilities. We use \textbf{Z3 (version 4.12.2)} as the SMT solver. Table~\ref{tab:hyperparams} details the specific configuration used across all experiments.

\begin{table}[h!]
\centering
    \begin{tabular}{l|c|}
        \toprule
        \textbf{Parameter} & \textbf{Value}  \\
        \midrule
        \multicolumn{2}{l}{\textit{LLM Generation}} \\
        Temperature ($T_{gen}$) & 1.0\\
        Top-P & 0.99 \\
        Max Output Tokens & 10,000  \\
        Thinking Budget & 4,000  \\
        \midrule
        \multicolumn{2}{l}{\textit{\forge Pipeline}} \\
        Max Iterations ($T_{max}$) & 3  \\
        Consensus Samples ($K$) & 3 \\
        Consensus Threshold & 0.70  \\
        Soft-Verify Judges ($N$) & 5  \\
        Acceptance Score ($\tau_{acc}$) & 0.75  \\
        Convergence $\epsilon$ & 0.01 \\
        Judge Model & Sonnet-3.7\\
        \bottomrule
    \end{tabular}
    \caption{Hyperparameters for the \forge pipeline. These values were held constant across all datasets (FOLIO, ZebraLogic, etc.) to demonstrate robustness.}
    \label{tab:hyperparams}
\end{table}

\subsection{Prompt Templates}
We employ a modular prompting strategy. The system prompts for the key components of the pipeline are provided below.

\subsubsection{Claim Decomposition \& Classification}
This prompt breaks the raw generation into atomic units and routes them to the appropriate verifier.

\begin{tcolorbox}[colback=gray!10, boxrule=0.5pt, title=System Prompt: Decomposition]
\small
\texttt{You are an expert logician. Analyze the provided Answer and decompose it into atomic, verifiable claims. For each claim, assign a Semantic Type based on the following definitions:}
\begin{itemize}[nosep]
    \item \texttt{MATHEMATICAL}: Involves arithmetic, algebra, or numerical constraints.
    \item \texttt{LOGICAL}: Involves first-order logic, set theory, or strict deduction.
    \item \texttt{COMMONSENSE}: Involves general world knowledge or vague predicates.
\end{itemize}
\texttt{Output format: JSON list of objects \{``text": string,``type": string\}.}
\end{tcolorbox}

\subsubsection{Autoformalization (Natural Language \texorpdfstring{$\rightarrow$}~SMT-LIB2)}
This prompt translates text into code. Note the instruction \texttt{(set-logic ALL)}, which allows the solver to handle mixed integer arithmetic and uninterpreted functions dynamically.

\begin{tcolorbox}[colback=gray!10, boxrule=0.5pt, title=System Prompt: Formalization]
\small
\texttt{Translate the following natural language context and claims into SMT-LIB2 format for the Z3 solver.}

\textbf{Constraints:}
\begin{enumerate}[nosep]
    \item \texttt{Declare all sorts (e.g., Person, Object) explicitly.}
    \item \texttt{Use (set-logic ALL) to support mixed arithmetic and logic.}
    \item \texttt{Do not include (check-sat) commands; simply assert the facts.}
    \item \texttt{If a statement is ambiguous, use uninterpreted functions.}
\end{enumerate}
\texttt{Input Context: [CONTEXT]} \\
\texttt{Input Claim: [CLAIM]} \\
\texttt{Output: <smt> ... code ... </smt>}
\end{tcolorbox}

\subsubsection{Feedback Injection (Refinement Step)}
When the SMT solver returns \texttt{UNSAT}, we calculate the Minimal Correction Set (MCS) and feed it back to the model using this template.

\begin{tcolorbox}[colback=gray!10, boxrule=0.5pt, title=System Prompt: Refinement]
\small
\texttt{Your previous answer contained logical contradictions verified by a formal solver. Please refine your reasoning.}
\textbf{Verification Report:}
\begin{itemize}[nosep]
    \item \texttt{Status: UNSATISFIABLE}
    \item \texttt{Conflict Logic: [Z3\_UNSAT\_CORE]}
    \item \texttt{Actionable Feedback: The set of claims \{C1, C3\} cannot be true simultaneously. Specifically, [MCS\_DESCRIPTION].}
\end{itemize}
\texttt{Instruction: Rewrite your answer to resolve this contradiction. Do not weaken the premises; correct your derivation.}
\end{tcolorbox}

\subsection{Consensus and Scoring Metrics}
\label{sec:appendix_metrics}

\paragraph{Formalization Consensus.}
To ensure the SMT code faithfully represents the natural language, we generate $K=3$ candidate formalizations for every claim. We compute the \textbf{Semantic Equivalence} between candidates by querying the solver for the unsatisfiability of their negation (see Eq.~\ref{eq:equivalence}). We construct an equivalence graph where edges represent proven logical identity. If a clique of size $\ge \lceil K/2 \rceil$ (majority consensus) is found, the representative formalization is accepted. Otherwise, the claim is flagged as ``Ambiguous'' and routed to Soft Verification.

\paragraph{Variance-Based Scoring.}
The final confidence score $\mathcal{S}(\mathcal{A})$ is penalized if high-confidence claims contradict each other. We define the score as:
\begin{equation}
    \mathcal{S}(\mathcal{A}) = \mu_{claims} \cdot \max\left(0.5, 1.0 - \frac{\sigma_{claims}}{\mu_{claims} + \epsilon}\right)
\end{equation}
where $\mu_{claims}$ is the weighted average verification status (Entailed=$1.0$, Soft-Pass=$0.9$, Possible=$0.7$, Fail=$0.0$) and $\sigma_{claims}$ is the standard deviation. This penalizes answers that contain a mix of ``True'' and ``False'' claims, encouraging internal consistency.

\subsection{Compute Infrastructure}
All experiments were conducted on a standard workstation (64GB RAM, 16-core CPU). The Z3 solver component handles both verification (avg. $<200$ms) and equivalence checking (capped at 2.0s timeout per pair). The LLM inference was performed using the AWS Bedrock API.

\subsection{Benchmark Descriptions}
We used test split of each dataset mentioned below:

\textbf{FOLIO} \citep{han2022folio}: First-order logic reasoning requiring entailment verification. Metric: Accuracy on predicting "Entailed/Contradictory/Unknown."

\textbf{ProofWriter} \citep{tafjord2021proofwriter}: Synthetic deductive reasoning. Metric: Proof step accuracy.

\textbf{ZebraLogic} \citep{lin2025zebralogic}: Constraint satisfaction puzzles (Einstein's riddle variants). Metric: Exact answer match.

\textbf{AR-LSAT} \citep{zhong2021ar}: Analytical reasoning from LSAT exams. Metric: Multiple-choice accuracy.

\textbf{BBEH} \citep{kazemi2025big}: Big-Bench Extra Hard reasoning subset. Metric: Task-specific accuracy (varies by sub-task).

\textbf{HLE} \citep{phan2025humanity}: Humanity's Last Exam—diverse reasoning including ethics, physics, literature (1,200 questions). Metric: Multiple-choice accuracy.

All reported numbers are \textbf{task accuracy} against gold labels, not internal verification scores. $S(\mathcal{A})$ is used only to decide whether to accept or refine during iteration.

\onecolumn
\section{Detailed Verification Trajectories}
\label{appendix:case study}
\subsection*{Case Study 1: AR-LSAT (Temporal Scheduling)}
\label{appendix:arlsat}

\begin{tcolorbox}[enhanced, colback=infobg, colframe=infoblue, title={\textbf{Problem Context \& Question}}, boxrule=1.5pt]
\small
\textbf{Context:} Of eight students—George, Helen, Irving, Kyle, Lenore, Nina, Olivia, and Robert—in a seminar, exactly six will give individual oral reports during three consecutive days—Monday, Tuesday, and Wednesday. Exactly two reports will be given each day—one in the morning and one in the afternoon—according to the following conditions:
\begin{itemize}[nosep]
    \item Tuesday is the only day on which George can give a report.
    \item \textbf{Neither Olivia nor Robert can give an afternoon report.}
    \item If Nina gives a report, then on the next day Helen and Irving must both give reports...
\end{itemize}
\textbf{Question:} If Kyle and Lenore do not give reports, then the morning reports on Monday, Tuesday, and Wednesday, respectively, could be given by: \\
\textbf{Options:} A) Helen, George, Nina \quad B) Irving, Robert, Helen \quad C) Nina, Helen, Olivia \quad D) Olivia, Robert, Irving \quad E) Robert, George, Helen
\end{tcolorbox}

\begin{tcolorbox}[skin=enhanced, colback=white, colframe=failred, title={\Large\bfseries Iteration 1 (Failed)}, boxrule=2pt, breakable]

\subsubsection*{Generated Answer}
\begin{tcolorbox}[colback=failbg, colframe=failred, sharp corners=west]
\textbf{Answer: E) Robert, George, and Helen}
\end{tcolorbox}

\subsubsection*{Extracted Claims \& Verification Status}
\scriptsize
\begin{tabularx}{\linewidth}{|c|L|c|c|c|Y|}
\hline
\rowcolor{darkgray} \textcolor{white}{\textbf{Idx}} & \textcolor{white}{\textbf{Claim Text}} & \textcolor{white}{\textbf{Type}} & \textcolor{white}{\textbf{Status}} & \textcolor{white}{\textbf{Score}} & \textcolor{white}{\textbf{Conf}} \\
\hline
0 & The selected answer is Option E & LOGICAL & \statusbadge{failred}{CONTRADICTORY} & 0.00 & 0.95 \\
1 & Robert gives the Monday Morning report & TEMPORAL & \statusbadge{passgreen}{POSSIBLE} & 0.70 & 0.98 \\
2 & George gives the Tuesday Morning report & TEMPORAL & \statusbadge{passgreen}{POSSIBLE} & 0.70 & 0.95 \\
3 & Helen gives the Wednesday Morning report & TEMPORAL & \statusbadge{passgreen}{POSSIBLE} & 0.70 & 0.92 \\
4 & Olivia gives an afternoon report (Implicit) & LOGICAL & \statusbadge{failred}{CONTRADICTORY} & 0.00 & 1.00 \\
\hline
\end{tabularx}

\subsubsection*{Detailed Failure Analysis}
\begin{tcolorbox}[colback=failbg, colframe=failred, title={\textbf{Claim 0 Failure: Constraint Violation}}, boxrule=0.5pt]
\textbf{Issue:} Option E assigns slots (Robert, George, Helen) to mornings. Since Kyle and Lenore are out, the remaining slots (Afternoons) must be filled by {Nina, Irving, Olivia}. This forces \textbf{Olivia} into an afternoon slot.
\begin{lstlisting}[style=smtstyle]
; Context Constraint: Neither Olivia nor Robert can give afternoon reports
(assert (not (exists ((d Day)) (gives_report Olivia d Afternoon))))

; Claim 0 implies:
(assert (gives_report Robert Monday Morning))
(assert (gives_report George Tuesday Morning))
(assert (gives_report Helen Wednesday Morning))
; This leaves Olivia with only Afternoon slots available -> CONTRADICTION
\end{lstlisting}
\textbf{Z3 Result:} \texttt{unsat} \\
\textbf{Unsat Core:} \texttt{['context\_axiom\_olivia', 'claim\_0 part\ 0']} \\
\textbf{Minimal Correction Subset (MCS):} Remove \texttt{(gives\_report Robert Monday Morning)} OR Remove \texttt{Option E}.
\end{tcolorbox}

\subsubsection*{Metrics (Iteration 1)}
\begin{tabularx}{\linewidth}{X X X X}
\textbf{Overall Score:} \statusbadge{failred}{0.350} & \textbf{Joint Cons:} 0.0\% & \textbf{Entailment:} 0.42 & \textbf{Verified:} False
\end{tabularx}
\end{tcolorbox}

\begin{center}
\huge $\Downarrow$ \normalsize \textbf{\forge Feedback: "Constraint Violation: Olivia forced into Afternoon. MCS: Change Option E or Robert's slot."} \huge $\Downarrow$
\end{center}

\begin{tcolorbox}[skin=enhanced, colback=white, colframe=passgreen, title={\Large\bfseries Iteration 2 (Success)}, boxrule=2pt, breakable]

\subsubsection*{Refined Answer}
\begin{tcolorbox}[colback=passbg, colframe=passgreen, sharp corners=west]
\textbf{Answer: D) Olivia, Robert, and Irving} \\
\textit{Reasoning:} The previous attempt failed because it forced Olivia into an afternoon slot. To satisfy the condition "Neither Olivia nor Robert can give an afternoon report," both must be assigned to morning slots. Option D places Olivia (Mon AM), Robert (Tue AM), and Irving (Wed AM). This leaves valid afternoon slots for the others.
\end{tcolorbox}

\subsubsection*{Extracted Claims \& Verification Status}
\scriptsize
\begin{tabularx}{\linewidth}{|c|L|c|c|c|Y|}
\hline
\rowcolor{darkgray} \textcolor{white}{\textbf{Idx}} & \textcolor{white}{\textbf{Claim Text}} & \textcolor{white}{\textbf{Type}} & \textcolor{white}{\textbf{Status}} & \textcolor{white}{\textbf{Score}} & \textcolor{white}{\textbf{Conf}} \\
\hline
0 & The selected answer is Option D & LOGICAL & \statusbadge{passgreen}{ENTAILED} & 1.00 & 0.95 \\
1 & Olivia gives the Monday Morning report & TEMPORAL & \statusbadge{passgreen}{POSSIBLE} & 0.95 & 0.99 \\
2 & Robert gives the Tuesday Morning report & TEMPORAL & \statusbadge{passgreen}{POSSIBLE} & 0.95 & 0.99 \\
3 & Irving gives the Wednesday Morning report & TEMPORAL & \statusbadge{passgreen}{POSSIBLE} & 0.95 & 0.99 \\
4 & Constraint on Olivia/Robert is satisfied & LOGICAL & \statusbadge{passgreen}{ENTAILED} & 1.00 & 1.00 \\
\hline
\end{tabularx}

\subsubsection*{Verification Success}
\begin{tcolorbox}[colback=passbg, colframe=passgreen, title={\textbf{Joint Consistency Verified}}, boxrule=0.5pt]
\begin{lstlisting}[style=smtstyle]
(assert (gives_report Olivia Monday Morning))
(assert (gives_report Robert Tuesday Morning))
(assert (gives_report Irving Wednesday Morning))
\end{lstlisting}
\textbf{Z3 Result:} \texttt{sat}. \\
\textbf{Entailment Check:} The conjunction of these assignments is consistent with all global constraints.
\end{tcolorbox}

\subsubsection*{Final Metrics}
\begin{tabularx}{\linewidth}{X X X X}
\textbf{Overall Score:} \statusbadge{passgreen}{0.965} & \textbf{Joint Cons:} 100\% & \textbf{Entailment:} 0.98 & \textbf{Verified:} \statusbadge{passgreen}{TRUE}
\end{tabularx}
\end{tcolorbox}

\newpage

\subsection*{Case Study 2: FOLIO (First-Order Logic)}
\label{appendix:folio}

\begin{tcolorbox}[enhanced, colback=infobg, colframe=infoblue, title={\textbf{Problem Context \& Question}}, boxrule=1.5pt]
\small
\textbf{Context:}
\begin{enumerate}[nosep]
    \item All guests at the party drink wine or beer. ($\forall x. Guest(x) => Wine(x) \lor Beer(x)$)
    \item If a guest drinks wine, they eat cheese. ($\forall x. Guest(x) \land Wine(x) => Cheese(x)$)
    \item \textbf{No one who eats cheese likes fish.} ($\forall x. Cheese(x) => \neg LikesFish(x)$)
    \item Alan is a guest. ($Guest(Alan)$)
    \item \textbf{Alan likes fish.} ($LikesFish(Alan)$)
\end{enumerate}
\textbf{Question:} Does Alan drink beer?
\end{tcolorbox}

\begin{tcolorbox}[skin=enhanced, colback=white, colframe=failred, title={\Large\bfseries Iteration 1 (Hallucination)}, boxrule=2pt, breakable]

\subsubsection*{Generated Answer}
\begin{tcolorbox}[colback=failbg, colframe=failred, sharp corners=west]
No, Alan drinks wine. Since Alan is a guest, he drinks wine or beer. Guests usually drink wine with meals, and there is no rule against it.
\end{tcolorbox}

\subsubsection*{Detailed Failure Analysis}
\begin{tcolorbox}[colback=failbg, colframe=failred, title={\textbf{Claim 0: "Alan drinks wine" is Contradictory}}, boxrule=0.5pt]
\textbf{SMT Formalization:}
\begin{lstlisting}[style=smtstyle]
(assert (Drinks Alan Wine))
\end{lstlisting}
\textbf{Constraint Propagation:}
1. \texttt{(Drinks Alan Wine)} $=>$ \texttt{(Eats Alan Cheese)} [Context Axiom 2]
2. \texttt{(Eats Alan Cheese)} $=>$ \texttt{(Not (Likes Alan Fish))} [Context Axiom 3]
3. Context Axiom 5 asserts: \texttt{(Likes Alan Fish)}
4. \textbf{Contradiction:} \texttt{(Likes Alan Fish)} AND \texttt{(Not (Likes Alan Fish))}

\textbf{Z3 Result:} \texttt{unsat}. \\
\textbf{Unsat Core:} \texttt{['Drinks\_Alan\_Wine', 'Axiom\_2', 'Axiom\_3', 'Axiom\_5']}
\end{tcolorbox}

\subsubsection*{Metrics (Iteration 1)}
\begin{tabularx}{\linewidth}{X X X X}
\textbf{Score:} \statusbadge{failred}{0.000} & \textbf{Joint Cons:} 0.0\% & \textbf{Verified:} False
\end{tabularx}
\end{tcolorbox}

\begin{center}
\huge $\Downarrow$ \normalsize \textbf{\forge Feedback: "Claim 'Alan drinks wine' causes contradiction with 'Alan likes fish'. MCS: Delete claim."} \huge $\Downarrow$
\end{center}

\begin{tcolorbox}[skin=enhanced, colback=white, colframe=passgreen, title={\Large\bfseries Iteration 2 (Success)}, boxrule=2pt, breakable]

\subsubsection*{Refined Answer}
\begin{tcolorbox}[colback=passbg, colframe=passgreen, sharp corners=west]
Yes, Alan drinks beer.
\textit{Reasoning:} We know Alan is a guest, so he drinks Wine or Beer.
If he drank Wine, he would eat Cheese. If he ate Cheese, he would not like Fish.
However, we know Alan DOES like Fish. Therefore, he cannot drink Wine.
Since he must drink Wine or Beer, and Wine is impossible, he must drink Beer.
\end{tcolorbox}

\subsubsection*{Extracted Claims \& Verification Status}
\scriptsize
\begin{tabularx}{\linewidth}{|c|L|c|c|c|Y|}
\hline
\rowcolor{darkgray} \textcolor{white}{\textbf{Idx}} & \textcolor{white}{\textbf{Claim Text}} & \textcolor{white}{\textbf{Type}} & \textcolor{white}{\textbf{Status}} & \textcolor{white}{\textbf{Score}} & \textcolor{white}{\textbf{Conf}} \\
\hline
0 & Alan drinks beer & LOGICAL & \statusbadge{passgreen}{ENTAILED} & 1.00 & 0.98 \\
1 & Alan does not drink wine & LOGICAL & \statusbadge{passgreen}{ENTAILED} & 1.00 & 0.98 \\
\hline
\end{tabularx}

\subsubsection*{Verification Success}
\begin{tcolorbox}[colback=passbg, colframe=passgreen, title={\textbf{Entailment Proof}}, boxrule=0.5pt]
\begin{lstlisting}[style=smtstyle]
; To prove "Alan drinks Beer" is entailed, we assert its negation and check for UNSAT
(push)
(assert (not (Drinks Alan Beer)))
(check-sat) ; Returns UNSAT -> Proves Entailment
(pop)
\end{lstlisting}
\end{tcolorbox}

\subsubsection*{Final Metrics}
\begin{tabularx}{\linewidth}{X X X X}
\textbf{Score:} \statusbadge{passgreen}{1.000} & \textbf{Joint Cons:} 100\% & \textbf{Entailment:} 1.00 & \textbf{Verified:} \statusbadge{passgreen}{TRUE}
\end{tabularx}
\end{tcolorbox}

\newpage

\subsection*{Case Study 3: ZebraLogic (Spatial Relations)}
\label{appendix:zebra}

\begin{tcolorbox}[enhanced, colback=infobg, colframe=infoblue, title={\textbf{Problem Context \& Question}}, boxrule=1.5pt]
\small
\textbf{Context:} Three houses (Red, Blue, Green) are in a row, numbered 1 (left) to 3 (right).
\begin{itemize}[nosep]
    \item The Blue house is immediately to the right of the Red house.
    \item The Green house is somewhere to the left of the Blue house.
\end{itemize}
\textbf{Question:} What is the order of the houses from left to right?
\end{tcolorbox}

\begin{tcolorbox}[skin=enhanced, colback=white, colframe=failred, title={\Large\bfseries Iteration 1 (Inconsistent)}, boxrule=2pt, breakable]

\subsubsection*{Generated Answer}
\begin{tcolorbox}[colback=failbg, colframe=failred, sharp corners=west]
The order is Red, Blue, Green.
\textit{Reasoning:} Red is at 1. Blue is right of Red, so Blue is 2. Green is left of Blue, so Green is 3... wait.
\end{tcolorbox}

\subsubsection*{Extracted Claims \& Verification Status}
\scriptsize
\begin{tabularx}{\linewidth}{|c|L|c|c|c|Y|}
\hline
\rowcolor{darkgray} \textcolor{white}{\textbf{Idx}} & \textcolor{white}{\textbf{Claim Text}} & \textcolor{white}{\textbf{Type}} & \textcolor{white}{\textbf{Status}} & \textcolor{white}{\textbf{Score}} & \textcolor{white}{\textbf{Conf}} \\
\hline
0 & Red is at position 1 & LOGICAL & \statusbadge{passgreen}{POSSIBLE} & 0.70 & 0.90 \\
1 & Blue is at position 2 & LOGICAL & \statusbadge{passgreen}{POSSIBLE} & 0.70 & 0.90 \\
2 & Green is at position 3 & LOGICAL & \statusbadge{failred}{CONTRADICTORY} & 0.00 & 0.90 \\
\hline
\end{tabularx}

\subsubsection*{Detailed Failure Analysis}
\begin{tcolorbox}[colback=failbg, colframe=failred, title={\textbf{Joint Inconsistency}}, boxrule=0.5pt]
\textbf{Issue:} Claim 2 (Green=3) contradicts Claim 1 (Blue=2) given the constraint "Green is left of Blue".
\begin{lstlisting}[style=smtstyle]
(declare-fun pos (House) Int)
(assert (= (pos Blue) (+ (pos Red) 1)))  ; Blue right of Red
(assert (< (pos Green) (pos Blue)))      ; Green left of Blue

; Claims:
(assert (= (pos Blue) 2))
(assert (= (pos Green) 3))

; Check: 3 < 2 ? -> FALSE.
\end{lstlisting}
\textbf{Z3 Result:} \texttt{unsat}. \\
\textbf{Unsat Core:} \texttt{['Green\_Left\_Blue\_Constraint', 'Claim\_1', 'Claim\_2']}
\end{tcolorbox}

\subsubsection*{Metrics (Iteration 1)}
\begin{tabularx}{\linewidth}{X X X X}
\textbf{Score:} \statusbadge{failred}{0.466} & \textbf{Joint Cons:} 0.0\% & \textbf{Verified:} False
\end{tabularx}
\end{tcolorbox}

\begin{center}
\huge $\Downarrow$ \normalsize \textbf{\forge Feedback: "Constraint violation: Green must be < Blue. Current: 3 < 2 (False)."} \huge $\Downarrow$
\end{center}

\begin{tcolorbox}[skin=enhanced, colback=white, colframe=passgreen, title={\Large\bfseries Iteration 2 (Success)}, boxrule=2pt, breakable]

\subsubsection*{Refined Answer}
\begin{tcolorbox}[colback=passbg, colframe=passgreen, sharp corners=west]
The order is Green, Red, Blue.
\textit{Reasoning:}
1. Blue is right of Red -> $B = R + 1$.
2. Green is left of Blue -> $G < B$.
Possible values: If R=1, B=2. Then G < 2 implies G=1. But houses are distinct, so G cannot be 1.
Therefore, Red cannot be 1.
Try R=2, B=3. Then G < 3 implies G=1 or G=2. Since R=2, G must be 1.
Solution: Green(1), Red(2), Blue(3).
\end{tcolorbox}

\subsubsection*{Extracted Claims \& Verification Status}
\scriptsize
\begin{tabularx}{\linewidth}{|c|L|c|c|c|Y|}
\hline
\rowcolor{darkgray} \textcolor{white}{\textbf{Idx}} & \textcolor{white}{\textbf{Claim Text}} & \textcolor{white}{\textbf{Type}} & \textcolor{white}{\textbf{Status}} & \textcolor{white}{\textbf{Score}} & \textcolor{white}{\textbf{Conf}} \\
\hline
0 & Green is at position 1 & LOGICAL & \statusbadge{passgreen}{ENTAILED} & 1.00 & 0.95 \\
1 & Red is at position 2 & LOGICAL & \statusbadge{passgreen}{ENTAILED} & 1.00 & 0.95 \\
2 & Blue is at position 3 & LOGICAL & \statusbadge{passgreen}{ENTAILED} & 1.00 & 0.95 \\
\hline
\end{tabularx}

\subsubsection*{Verification Success}
\begin{tcolorbox}[colback=passbg, colframe=passgreen, title={\textbf{Entailment Verified}}, boxrule=0.5pt]
\begin{lstlisting}[style=smtstyle]
(assert (= (pos Green) 1))
(assert (= (pos Red) 2))
(assert (= (pos Blue) 3))
\end{lstlisting}
\textbf{Z3 Result:} \texttt{sat}. The solution uniquely satisfies all spatial constraints.
\end{tcolorbox}

\subsubsection*{Final Metrics}
\begin{tabularx}{\linewidth}{X X X X}
\textbf{Score:} \statusbadge{passgreen}{1.000} & \textbf{Joint Cons:} 100\% & \textbf{Entailment:} 1.00 & \textbf{Verified:} \statusbadge{passgreen}{TRUE}
\end{tabularx}
\end{tcolorbox}

\end{document}